\documentclass[preprint,12pt]{elsarticle}

\newcommand{\nummodels}{32} 

\newcommand{\gptfivetwo}{\emph{GPT-5.2}}
\newcommand{\bestmodel}{\emph{GPT-5.2}}
\newcommand{\gptthreefive}{\emph{GPT-3.5 Turbo}}
\newcommand{\kimi}{\emph{Kimi K2}}

\newcommand{\qstrlarge}{\emph{QSTRBenchExtended}}
\newcommand{\qstrsmall}{\emph{QSTRBench}}

\newcommand{\repeats}[4]{(#1:#2:#3/#4)} 

\usepackage{float}
\usepackage{booktabs}
\usepackage[hidelinks]{hyperref}
\usepackage{xurl}
\usepackage{array}
\usepackage{makecell}
\usepackage{tablefootnote}
\usepackage{tikz}
\usepackage{tabularx}
\usetikzlibrary{arrows.meta, positioning, calc}
\usepackage{listings}
\usepackage{longtable}
\usepackage{enumitem}
\setlist{nolistsep}

\NewDocumentCommand{\relation}{m}{\ensuremath{\mathsf{#1}}}

\newcommand{\pr}[1]{\textsf{#1}}
\renewcommand{\P}[0]{\pr{P}}

\newcommand{\PO}[0]{\pr{PO}}
\newcommand{\TPP}[0]{\pr{TPP}}

\newcommand{\DC}[0]{\pr{DC}}
\newcommand{\PP}[0]{\pr{PP}}
\newcommand{\EQ}[0]{\pr{EQ}}
\newcommand{\EC}[0]{\pr{EC}}
\newcommand{\NTPP}[0]{\pr{NTPP}}
\newcommand{\NTPPi}[0]{\pr{NTPPi}}
\newcommand{\TPPi}[0]{\pr{TPPi}}
\newcommand{\PPi}[0]{\pr{PPi}}
\newcommand{\PPI}[0]{\pr{PPi}}

\newcommand{\DR}[0]{\pr{DR}}

\newcommand{\IOD}[0]{\pr{IOD}}
\newcommand{\IOE}[0]{\pr{IOE}}
\newcommand{\IPD}[0]{\pr{IPD}}
\newcommand{\IPE}[0]{\pr{IPE}}
\newcommand{\OID}[0]{\pr{OID}}
\newcommand{\OIE}[0]{\pr{OIE}}
\newcommand{\OOD}[0]{\pr{OOD}}
\newcommand{\OOE}[0]{\pr{OOE}}
\newcommand{\OPD}[0]{\pr{OPD}}
\newcommand{\OPE}[0]{\pr{OPE}}
\newcommand{\PID}[0]{\pr{PID}}
\newcommand{\PIE}[0]{\pr{PIE}}
\newcommand{\POD}[0]{\pr{POD}}
\newcommand{\POE}[0]{\pr{POE}}
\newcommand{\PPD}[0]{\pr{PPD}}
\newcommand{\PPE}[0]{\pr{PPE}}

\usepackage{amssymb}
\usepackage{amsmath}

\journal{Artificial Intelligence}

\begin{document}

\begin{frontmatter}




\title{\qstrsmall: a New Benchmark to Evaluate the Ability of Language Models to Reason with Qualitative Spatial and Temporal Calculi} 




\author[1,2,3]{A.G. Cohn}

\author[1]{R.E. Blackwell}

\affiliation[1]{organization={School of Computer Science, The~University~of~Leeds},
                addressline={Woodhouse~Lane},
                city={Leeds},
                postcode={LS2~9JT},
                postcodesep={,},
                country={UK}}

\affiliation[2]{organization={The Alan Turing Institute},
                addressline={British Library, 96 Euston Road},
                city={London},
                postcode={NW1~2DB},
                postcodesep={,},
                country={UK}}

\affiliation[3]{organization={Tongji University},
                city={Shanghai},
                country={China}}
\begin{abstract}
We introduce an
extensive
qualitative spatial and temporal
reasoning (QSTR) benchmark for evaluating large language models
(LLMs). We pose questions concerning compositional reasoning (using
composition tables, CT), converse relations, and conceptual
neighbourhoods (CN) for QSTR calculi, 
Point Algebra (PA), Allen's Interval
Algebra,  Interval and Duration (INDU), Region Connection Calculus (RCC-5, RCC-8,
and RCC-22), the nine intersection model, cardinal direction calculus, and
STAR. The RCC-22 CN is published here for the first
time.  An extended benchmark systematically varies question
presentation including prefix/infix, words/symbols/nonce terms and
schematic descriptions for selected calculi.  We report results for contemporary frontier
models. All models tested perform better than guessing but none can
consistently answer all questions correctly.  Performance varies sharply by
calculus, with PA being the most straightforward, and RCC-22 the most
difficult.
We release the benchmark, and our results
under an open licence to facilitate further assessment of qualitative
spatio/temporal reasoning in LLMs.

\end{abstract}




\end{frontmatter}



\section{Introduction}

Qualitative Spatial and Temporal Reasoning (QSTR\footnote{We may use QSTR as shorthand for both Qualitative Spatial and Temporal Reasoning and Qualitative Spatial and Temporal Representation(s); context should  make clear which is intended.}) \cite{cohn2008qualitative,chen2015survey,cohn2001qualitative} is a well developed field which is concerned with the representation of qualitative spatial and temporal information and reasoning with it.  In natural language, spatial and temporal information is usually represented qualitatively  (using prepositions such as \emph{on, in, left of, part of, under, touching, before, after, ...)} and many calculi have been developed to represent such information {\cite{dylla2017}.

\emph{Large Language Models} (LLMs) \cite{devlin-etal-2019-bert,brown2020language}, such as
GPT, Gemini, Grok, Claude and Llama
are examples of so called \emph{Foundation Models} \cite{bommasani2021opportunities} which have been trained on very large textual corpora in order to generate text in response to a prompt. This is not the place to survey this burgeoning field, but we note that many claims have been made about the power and apparent intelligent behaviour that these models can display.  In particular, their performance on some benchmarks may lead one to believe that they possess
well developed reasoning capabilities. Indeed, it has been claimed that LLMs are doubling in performance every seven months \cite{kwa2025measuringaiabilitycomplete}.  So the question arises as to whether
LLMs can perform the reasoning commonly associated with qualitative spatial and temporal calculi.

This paper addresses this question. We create a benchmark, \qstrsmall, which is designed to test three kinds of reasoning tasks commonly associated with QSTR calculi: what is a relation's converse, what is the composition of two relations, and what are the conceptual neighbours of a relation. Since LLMs are well known to be sensitive to the precise phrasing of a question, the benchmark poses these questions in multiple ways, such as the use of infix and prefix notations, and English words versus symbols. Since LLMs are widely believed not to genuinely reason but rather simply replicate patterns in the training data, we also have variants of the prompts where the relation names are anonymised. An extended benchmark, \qstrlarge, takes this further by including prompts which are intended to probe this issue further, for example by using a schematic prompt where the relations are not described using text but illustrated using simple \emph{ASCII art} diagrams, and by swapping relations names associated with the textual semantic descriptions. The benchmarks cover the three best known temporal qualitative calculi, and six of the best known qualitative spatial calculi. Having created the two benchmarks we then test the performance of 32 LLMs ranging from recent large commercial models to much smaller open weights models. Of course the LLM landscape is constantly changing so this evaluation is just a snapshot -- the  lasting value of the paper is the benchmark and its methodology.

The structure of the paper is as follows.  Section 2 covers related work, and section 3 the experimental design - the construction of the two benchmarks. Section 4 presents the results of the evaluation of the selected LLMs on the two benchmarks. Section 5 is a summary of the paper. The paper concludes with a discussion of its limitations in section 6 and future work in section 7.

\section{Related Work}

A QSTR calculus is a formal system comprising a (usually small) finite set of relation symbols, a syntax for combining them, and a set of axioms and inference rules for reasoning about them. Each symbol in the calculus denotes a qualitative binary relation over a specific domain (such as spatial regions or temporal intervals). The \emph{base} relations of the calculus comprise a jointly exhaustive and pairwise disjoint  (JEPD) set, with other relations in the calculus formed by disjunction over the base relations. In this paper, we focus on the base relations and reasoning over them. The calculus thereby defines a relation algebra, enabling formal reasoning about configurations of entities based on the relations they satisfy.

Relational composition is well studied from a theoretical point of view. R3($x,z$) is the composition of R1($x,y$) and R2($y,z$) if it is implied by the conjunction of the latter two relations.  In general, R3($x,z$) is a disjunction of relations.
 A \emph{composition table} (CT) records the results for all combinations of base relations in a particular calculus.

\begin{figure}
\centering
\includegraphics[width=0.6\columnwidth]{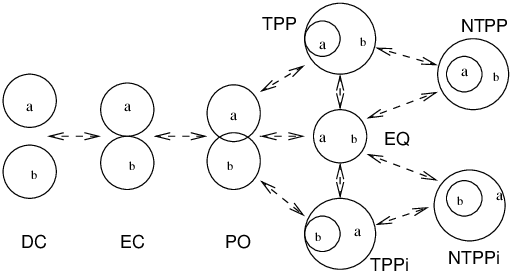} 
\caption{The eight base relations of  RCC-8  illustrated in 2D \protect\cite{cohn1997qualitative}:  \DC\ (Disconnected), \EC\ (Externally Connected), \PO\ (Partially Overlapping), \TPP\ (Tangential Proper Part), \NTPP\ (Nontangential Proper Part) and \EQ\ (Equals); \TPPi\  and \NTPPi\ are the converses of \TPP\  and \NTPP\  respectively since they are asymmetric. The arrows denote relations which are \emph{conceptual neighbours}. }
\label{rcc8-cn-diagram}
\end{figure}

A second form of reasoning commonly associated with QSTR involves the \emph{conceptual neighbourhood} (CN) \citep{freksa1992temporal} -- also called a \emph{continuity network}~\cite{randell1989modelling}. Fig. \ref{rcc8-cn-diagram} illustrates the CN for RCC-8 -- there is an edge between two relations iff one relation can be directly followed by the second one, assuming continuous translation or morphological change.

A third form of QSTR reasoning considers the \emph{converse} of a relation, which refers to the relation obtained by reversing the arguments; if \( R(x, y) \) holds, then its converse \( R^{\smile}(y, x) \) captures the same spatial configuration from the opposite perspective.  For example,  in RCC-8, \TPPi\  is the converse of \TPP, and \PO\ is its own converse.

The ability of LLMs to reason about RCC-8 was first explored by \citet{cohn2023evaluation} using ChatGPT-4.
A subsequent preprint  \citep{cohn2024largelanguagemodelsreason} extended this using  API interfaces to six different models, but omitted testing the EQ relation. \citet{gardelakos2025can} tested RCC-8 and the Cardinal Direction Calculus (CDC) showing that more recent large reasoning models, and context, increased RCC-8 reasoning accuracy, but that models showed inductive bias.

\citet{bellodi2025assessing} looked at the ability of LLMs to reason about temporal intervals, concluding that LLMs are unreliable when reasoning about unseen problems.
Fatemi et al \cite{fatemi2024test} introduced Test of Time (ToT), a benchmark for LLM temporal reasoning that uses synthetic graph-structured facts and anonymized WikiData. They found that  temporal reasoning  in the tested LLMs is highly sensitive to graph structure, graph size, and the ordering of facts presented to the LLM.

\citet{topsakal2025evaluating} looked at LLM performance in playing grid based games -- although these games require spatial awareness, spatial reasoning is not the focus of the work.
\citet{yamada2024evaluating} considered navigation as spatial reasoning, concluding that spatial knowledge is only partially learned from text and that while LLMs encode some implicit spatial structure, their reasoning remains inconsistent and error-prone.
The BLINK benchmark \cite{fu2024blink} includes spatial reasoning questions but these are geared towards perception of spatial arrangements within images presented to multimodal LLMs.

Yin et al
\cite{yin2025multimodal} tested Multimodal\footnote{There is also other work on testing spatial reasoning in a multi-modal setting, typically by testing whether such models can correctly infer spatial relations between objects in a presented scene, but that is not directly relevant to the purely linguistic tasks we consider here, so we do not survey this work.} Language Models (MLMs) with multiple choice
questions concerning compass directions and egocentric directions between
objects on a grid. They showed that MLMs do not understand compass
direction reasoning well, and often guess. The work predates the best frontier models tested here.
\citet{cohn2025evaluatingabilitylargelanguage} presented a benchmark for testing cardinal direction reasoning, but this uses informal questions and is not based on a rigorous calculus.
Xie et al \cite{xie2025evaluating} tested LLMs with RCC-8-like relations between US states and time zones, finding that while LLMs achieve better than chance performance, their reasoning is unreliable.

There are a few works which selectively test compositional reasoning, e.g. \citet{cohn2023dialectical} which investigated a number of spatial reasoning problems, including some limited instances of relational composition, but not exhaustively.
   Other work investigating the spatial reasoning abilities of LLMs which typically revolve around especially constructed  benchmarks such as Step\-Game \cite{li2024advancing,shi2022stepgame}   can also be regarded as testing compositional reasoning, but not in a methodical or exhaustive manner. StepGame aims to test an LLM's ability to correctly determine the qualitative direction relationship between two objects, given a set of direction relations between a larger set of objects, and between 1 and 10\footnote{Actually, as pointed out subsequently, the k=10 case in StepGame does not necessarily involve 10 steps, only up to 10 steps, so many instances are easier than when ten steps are truly necessary\cite{mcpheat2025decompsrdatasetdecomposedanalyses}. Morever, there are template errors in automated translation of  the ground truth to a controlled English representation \citep{li2024advancing}, which cast further difficulty in interpreting the original results\citep{shi2022stepgame}. } reasoning steps are required to correctly determine the result.
Not surprisingly, performance deteriorates as the required number of steps increases. Performance increases markedly when the LLM is used to translate from the English specification to a logical representation and symbolic reasoning is used to compute the relationship. A follow-on dataset, DecompSR \citep{mcpheat2025decompsrdatasetdecomposedanalyses} corrects some issues in the original StepGame dataset and extends it to much longer step lengths (up to 100, with the possibility to generate arbitrary length via the program supplied). Like StepGame, DecompSR only tests directions, and while the paper shows that the tested LLMs are relatively immune to linguistic variations (though Swedish is worse than either English or Hindi), all LLMs show degraded performance as the number of steps increases, whether the inference is entirely within the LLM, or whether an external symbolic reasoner is employed with the LLM merely acting as translator (the probability of correctly translating all the statements decreases as the number of steps/statements increases).

The SpartQA dataset \cite{mirzaee-etal-2021-spartqa} is also focused on assessing spatial reasoning, but does not test composition or conceptual neighbourhoods.  The bAbI dataset \cite{weston2016towards}  also has some tasks which test spatial reasoning and compositional reasoning, in particular about directions to a limited extent. Other work  has investigated whether LLMs can acquire an understanding of a spatial environment from a turn-by-turn description of a route, with landmarks named at each turn; whilst the LLMs did perform reasonably well, the experiment did not involve any mereotopological relations, only left/right and
up/down~\cite{yamada2024evaluating}.

Weaknesses in the reasoning powers of LLMs have previously been noted (e.g. \citet{cai2023human}) so one might not expect LLMs to perform well in this regard. But on the other hand, there are a large number of papers about QSTR in the literature. These are likely to have formed part of the  training corpus of an LLM, and thus might facilitate correctly responding to prompts --  though the information concerning the actual reasoning steps are often given in tables (in particular
\emph{CT}s), or as diagrams (in particular CNs)
and thus might be hard for LLM training procedures to process effectively.

\section{Experimental Design}
\label{sec:exp-design}

We aim to construct a
benchmark that extensively
tests LLM performance over a range of qualitative spatial and temporal reasoning calculi.
There are many QSTR calculi \citep{dylla2017} and so we choose well-known
examples where the CT, CN, and converse relations
are readily available (Table \ref{tab:calculi}). Although ternary calculi exist (e.g. \citep{isli2000new,freksa2005using}), we focus here on calculi with binary relations, postponing any ternary calculus to further work. We prefer calculi
available in open source tooling such as GQR \citep{gantner2008gqr} or SparQ \citep{wolter2009sparq} to avoid transcription errors.
PA, IA and INDU are the mostly widely known temporal calculi. RCC-8 is probably the best known qualitative spatial calculus. We also
test 9IM \citep{egenhofer19949} as an alternative formulation of the relations to be found in RCC-8. We test CDC \citep{frank1996qualitative} and STAR \citep{renz2004qualitative} because of our interest in cardinal directions \citep{cohn_et_al:LIPIcs.COSIT.2024.28,cohn2025evaluatingabilitylargelanguage}.
\begin{table}[!htbp]
\centering
\caption{The list of QSTR calculi that we test; $n$ is the number of base relations. The number of citations are according to Google Scholar, January 2026.}
\vspace{0.5em} 
\resizebox{\textwidth}{!}{%
\begin{tabular}{llrlr}
\toprule
\textbf{Calculus} & \textbf{Abbrev.} & \textbf{$n$} & \textbf{Reference} & \textbf{Citations\footnote{Citation counts were made at the point of benchmark construction.}} \\
\midrule
Allen's Interval Algebra & IA       & 13 & \cite{allen1983maintaining}  & 13017        \\
Region Connection Calculus 8         & RCC-8      & 8  & \cite{randell1992spatial}\tablefootnote{This paper won the 2020 KR Test-of-Time award.} \cite{cohn1997qualitative} & 3372 + 617 \\
Point Algebra & PA       & 3 & \cite{vilain1986constraint} & 1487 \\
Cardinal Direction Calculus         & CDC        & 9 & \cite{frank1996qualitative} & 628\\
Region Connection Calculus 5         & RCC-5      & 5  & \cite{bennett1994spatial, jonsson1997}  & 286 + 105    \\
Egenhofer's 9 Intersection Model       & 9IM      & 8  & \cite{egenhofer1994deriving} & 296 \\
Region Connection Calculus 22         & RCC-22     & 22 &  \cite{cui1993} & 257\\
Revised STAR & STAR & 9 & \cite{renz2004qualitative}& 194\\
Interval and duration algebra & INDU      & 25 & \cite{pujari1999indu} & 124\\
\bottomrule
\end{tabular}%
}
\label{tab:calculi}
\end{table}
We also considered and rejected\footnote{The cost of experiments on commercial LLMs was a factor limiting the number of possible calculi we could select.} the following SPARQ and GQR calculi:
\begin{itemize}
\item \emph{Dipole Algebra} \citep[209 citations]{schlieder1995reasoning}, \citep[192 citations]{moratz2011condensed}\footnote{Citation data correct at the time of writing this section.} because
even the coarsest Dipole algebra (DRAc) has 24 base relations and leads to repetitive questions.
\item \emph{Single Cross} and \emph{Double Cross} \citep[893 citations]{freksa2005using} because these are ternary calculi.
\item \emph{FlipFlop Calculus} \citep[132 citations]{ligozat1993qualitative}, again, a ternary calculus.
\item \emph{Oriented Point Relation Algebra (OPRA)} \citep[87 citations]{moratz2005relative} -- OPRA has 72 base relations so is very large and rather expensive to test.
\item \emph{Absolute Distance Calculus} \citep[469 citations]{clementini1997qualitative},  and Relative Distance Calculus \citep[314 citations]{hernandez1995qualitative} because the CTs are uninteresting.
\item \emph{OPRA star} \citep{moratz2010extending}:  very low citations and closely related to the STAR calculus which we do test.
\item \emph{Block Algebra} \citep[323 citations]{guesgen1989spatial}, \citep[69 citations]{balbiani2002tractability} because the CT and CN is uninteresting -- the calculus is just the 3-way cross product of the interval algebra.
\item \emph{Occlusion Calculus} \citep[34 citations]{kohler2002occlusion}: this calculus is essentially just the cross-product of RCC-5 and the PA.
\item \emph{Qualitative Trajectory Calculus (QTC)} \citep[118 citations]{van2005qualitative}. There are very many variants of QTC and it is not clear which one(s) to choose; moreover the shortest path between any two relations in the conceptual neighbour is typically only two (i.e. with only one intervening relation, the ``0'' relation).
\item \emph{Dependency Calculus} \citep[15 citations]{ragni2005dependency}: this calculus is closely related to RCC-5, with the same number of relations, and a tractability-preserving homomorphism to RCC-5.
\item \emph{Point Branching Calculus} \citep[52 citations]{broxvall2003point}. This calculus is closely related to PA, but has an additional relation for points which are not ordered. We did not include this calculus owing to this similarity.
\item \emph{RCC-23}: there is no CN given in the literature for RCC-23; moreover one of the relations, INSIDE\_INSIDE\_EC, has no corresponding DC version and can only be realised with regions which are not fully connected. Therefore,  we instead chose RCC-22 which omits this relation.
\item \emph{Geometric Orientation Calculus} \citep[17 citations]{dylla2008agent}.  We exclude this calculus because of relative low citation count.
\item \emph{Cyclic ordering calculi} \citep[102 citations]{isli2000new}, because many variants are ternary.
\end{itemize}

For each calculus, we test all combinations of relations R1 and R2 from the CT, using a question of the form
\emph{{If R1(x,y) and R2(y,z) then what are the possible relations between x and z?}}.
We test the converse of each relation R using a question of the form:
\emph{If R(x,y) then what is the relation between y and x?}.
We test the CN of a relation R using a question of the form:
\emph{What are the conceptual neighbours of R(x,y)?} (except for 9IM\footnote{The 9IM description does not talk about conceptual neighbourhoods but closest topological distances, measured in terms of the similarity of the matrices defining two relations; closest topological distance is related to but not identical to the notion of two relations being conceptual neighbours.} when we use the following
wording: \emph{The closest topological relationships of a relation R between entities x and y is the set of relations with the least non-zero topological distances from R ... What are the closest topological relationships to R(x,y)?}).

The RCC-22 calculus is derived from RCC-23 \cite{cohn1997qualitative} but with the INSIDE\_\-INSIDEi\_\-EC relationship removed, which as noted above does not have a corresponding INSIDE\_\-INSIDEi\_\-DC relationship, and can only be realised with regions which are either multi-piece regions, or only point-connected.

We have a total of 102 base relations across all the tested calculi. So there are 102 converse questions,
102 conceptual neighbourhood questions and 1602 composition table questions\footnote{We test only compositions   within a calculus, not across calculi, so the total number is not $102^2$.}, a total of 1806 questions in our \qstrsmall\ data set.

Our \qstrlarge\ data set poses the questions in infix form x R y as well as prefix form with symbolic, natural-language and nonce word\footnote{ ``A nonce word (from the 16th-century phrase for the nonce, meaning `for the once') is a lexeme created for temporary use, to solve an immediate problem of communication.'' The Cambridge encyclopedia of the English language \citep{crystal2018cambridge}. See later for more information.} relation names, so 6 x 1806 = 10836 questions. We  test eponymous and anonymous schematic descriptions\footnote{Explained further below.} for PA (30 questions), IA (390 questions), INDU (1350 questions), RCC-5 (70 questions), RCC-8 (160 questions), and RCC-22 (1056 questions). We test RCC-8 with a point based description eponymously in prefix notation  (8 base relations = 8 + 8 + 8 * 8 = 80 questions), formulaic description\footnote{i.e. defining all the RCC-8 relations using the first order theory represented as formulae and axioms\cite{randell1992spatial}.} in \LaTeX (80 questions), and with a formulaic description in Unicode (80 questions). We also test the eponymous prefix description of RCC-8 with the relation descriptions in reverse order (80 questions) and with the relation descriptions deleted altogether (80 questions)\footnote{This tests if the model can reason purely from material in its training data.}. We also try to mislead models by swapping RCC-8 relation names, \EQ\ with \EC, \DC\ with \PO, \TPP\ with \NTPPi, and \TPPi\ with \NTPP\ (80 questions).  The \qstrlarge\ benchmark size is therefore 14372 questions (10836 + 30 + 390 + 1350 + 70 + 160 + 1056  + 80 + 80 + 80 + 80 + 80 + 80). Examples of these prompts are given in \mbox{\ref{example-prompts}}.
The \qstrlarge\ benchmark is too expensive for us to test on all models, but we are grateful to Microsoft Research for funding that allowed us to test with o1 on Microsoft Azure.

\begin{table}[!htbp]
\caption{The Region Connection Calculus, RCC-22 conceptual neighbourhood.}
\vspace{0.5em} 
\label{tbl:rcc22cn}
\begin{tabularx}{\linewidth}{lX}
\toprule
Relation & Conceptual Neighbours \\
\midrule
\EQ\ & \PO, \TPP, \NTPP, \TPPi, \NTPPi \\
\IOD\ & \IOE, \IPD, \POD, \PPD \\
\IOE\ & \PO, \IOD, \IPD, \IPE, \POD, \POE, \PPD, \PPE \\
\IPD\ & \IOD, \IOE, \IPE, \PPD \\
\IPE\ & \PO, \IOE, \IPD, \PPD, \PPE \\
\NTPP\ & \EQ, \TPP \\
\NTPPi\ & \EQ, \TPPi \\
\OID & \OIE, \OPD, \PID, \PPD \\
\OIE\ & \PO, \OID, \OPD, \OPE, \PID, \PIE, \PPD, \PPE \\
\OOD\ & \OOE, \OPD, \POD, \PPD \\
\OOE\ & \PO, \OOD, \OPD, \OPE, \POD, \POE, \PPD, \PPE \\
\OPD\ & \OID, \OIE, \OOD, \OOE, \OPE, \PPD \\
\OPE\ & \PO, \OIE, \OOE, \OPD, \PPD, \PPE \\
\PID\ & \OID, \OIE, \PIE, \PPD \\
\PIE & \PO, \OIE, \PID, \PPD, \PPE \\
\PO & \EQ, \TPP, \TPPi, \OOE, \IPE, \IOE, \PIE, \PPE, \POE, \OIE, \OPE \\
\POD\ & \IOD, \IOE, \OOD, \OOE, \POE, \PPD \\
\POE\ & \PO, \IOE, \OOE, \POD, \PPD, \PPE \\
\PPD\ & \IOD, \IOE, \IPD, \IPE, \OID, \OIE, \OOD, \OOE, \OPD, \OPE, \PID, \PIE, \POD, \POE, \PPE \\
\PPE\ & \PO, \IOE, \IPE, \OIE, \OOE, \OPE, \PIE, \POE, \PPD \\
\TPP & \EQ, \PO, \NTPP \\
\TPPi & \EQ, \PO, \NTPPi \\
\bottomrule
\end{tabularx}
\end{table}

For INDU and 9IM, we use the CT and converse data from SparQ \citep{wolter2009sparq}. All other calculi use  GQR \citep{gantner2008gqr}, except
RCC-22 which is based on the RCC-23 support in GQR (we removed the IIC relation). All CN information was transcribed from the literature, except for RCC-22 which we determined and publish here for the first time (Table \ref{tbl:rcc22cn})\footnote{This was generated using a Prolog program which can be found in  \ref{sec:prolog-program}.}.

We note that relations are not consistently named within the literature, and
so rather than test all combinations, we prefer the names used in the original reference or the names  we perceive to be in most common use.

We pose all relation questions in terms of variables x and y (we do this for consistency across calculi, although we acknowledge that other variable names such as A and B are also used in the literature).

Where possible we order relation descriptions based on the ordering in the primary literature, except for the RCC-8 reverse experiment
where we reverse the order for comparison purposes.

In some literature, the base relations are defined in terms of pictorial representations. Because we are working with LLMs, we provide a textual description of each relation using natural language (in English\footnote{Exploring reasoning in other languages remains for future work.}). However, for
IA, INDU, PA, RCC-5, RCC-8, and RCC-22 we also test schematic descriptions where the relations are described in terms of textual diagrams.

Nonce words were generated by randomly sampling a Markov chain model using trigrams from Jane Austen's \emph{Pride
and Prejudice}. Words were discarded unless they had seven letters and a Levenshtein edit distance of at least two
from all words in the Hunspell English language dictionary\footnote{\url{https://github.com/hunspell/hunspell} -- last accessed April 2026.}.

When using nonce word relation names, and in the RCC-8 experiment where we swapped relation names, we replace the name of the calculus with ``The calculus of interest in this question''.

In each section below we describe the calculi, illustrating them pictorially\footnote{The figures are purely for exposition purposes in this paper -- the LLMs were not provided with any diagrammatic illustration of the relations, except in the schematic experiment mentioned above.}. We include excerpts from the prompts but the full details can be found in \ref{example-prompts}.

The models tested are shown in Table \ref{tbl:models}. We record the number of parameters where known,
but many vendors do not publish architectural details,
presumably for competitive reasons. We tried prompting the models themselves, and asking
them to disclose their parameter counts, without success. However, model performance is based on
a range of factors including training data size, training data quality, and training methodology \citep{chang2024scaling}.

\begin{table}[!htbp]
\centering
\caption{Table of the \nummodels{} LLMs tested. Date is the release date. Params in the number of parameters in billions (but note that many vendors no longer publish the number of parameters). Context is the context window size. Input and output prices are in US   dollars per million tokens as of January 2026 (most experiments were run well before this date). Where no pricing is shown, the model was either run locally or has been discontinued.}
\vspace{0.5em} 
\resizebox{\textwidth}{!}{%
\begin{tabular}{@{}l l r r r r@{}}
\toprule
Date & Model & Params & Context & Input price & Output price \\
\midrule
2025-12-10 & openai/gpt-5.2 & - & 400,000 & 1.75 & 14.00 \\
2025-12-01 & deepseek/deepseek-v3.2 & - & 163,840 & 0.25 & 0.40 \\
2025-11-17 & x-ai/grok-4.1-fast & - & 2.000,000 & 0.20 & 0.50 \\
2025-11-12 & openai/gpt-5.1 & - & 400,000 & 1.25 & 10.00 \\
2025-08-21 & deepseek/deepseek-chat-v3.1 & 685 & 163,840 & 0.56 & 1.68 \\
2025-08-07 & openai/gpt-5 & - & 400,000 & 1.25 & 10.00 \\
2025-08-5 & openai/gpt-oss-20b & 21 & 131,072 & 0.02 & 0.10 \\
2025-08-5 & openai/gpt-oss-120b & 117 & 131,072 & 0.04 & 0.19 \\
2025-07-25 & z-ai/glm-4.5 & 355 & 131,072 & 0.35 & 1.55 \\
2025-07-11 & moonshotai/kimi-k2 & 1000 & 63,000 & 0.50 & 2.40 \\
2025-07-09 & x-ai/grok-4 & - & 256,000 & 3.00 & 15.00 \\
2025-06-26 & google/gemini-2.5-pro & - & 1,048,576 & 1.25 & 10.00 \\
2025-06-26 & google/gemini-2.5-flash-lite & - & 1,048,576 & 0.07 & 0.30 \\
2025-06-26 & google/gemini-2.5-flash & - & 1,048,576 & 0.30 & 2.50 \\
2025-05-28 & deepseek/deepseek-r1-0528 & 685 & 163,840 & 0.45 & 2.15 \\
2025-05-14 & anthropic/claude-sonnet-4 & - & 200,000 & 3.00 & 15.00 \\
2025-04-18 & meta-llama/llama-3-8b-instruct & 8.03 & 8,192 &  &  \\
2025-04-16 & openai/o4-mini & - & 200,000 & 1.10 & 4.40 \\
2025-04-16 & openai/o3 & - & 200,000 & 2.00 & 8.00 \\
2025-03-17 & amazon/nova-pro-v1 & - & 300,000 & 0.80 & 3.20 \\
2025-03-10 & google/gemma3-1b & 1 & 32,000 &  &  \\
2025-02-25 & google/gemini-2.0-flash-001 & - & 1,000,000 & 0.10 & 0.40 \\
2025-02-19 & anthropic/claude-3.7-sonnet & - & 200,000 & 3.00 & 15.00 \\
2025-02-17 & x-ai/grok-3-mini & - & 131,072 & 0.30 & 0.50 \\
2025-02-07 & openai/gpt-45-preview & - &  &  &  \\
2025-01-20 & deepseek/deepseek-r1-distill-qwen-14b & 14 & 64,000 &  &  \\
2025-01-20 & deepseek/deepseek-r1 & 685 & 163,840 & 0.70 & 2.50 \\
2025-01-10 & microsoft/phi4-reasoning-14b & 14 & 16,384 &  &  \\
2024-12-17 & openai/o1 & - & 200,000 & 15.00 & 60.00 \\
2024-12-06 & meta-llama/llama-3.3-70b-instruct & 70.6 & 131,072 & 0.01 & 0.32 \\
2024-11-20 & openai/gpt-4o-2024-11-20 & - & 128,000 & 2.50 & 10.00 \\
2024-01-25 & openai/gpt-3.5-turbo & - & 16,385 &  &  \\
\bottomrule
\end{tabular}%
}
\label{tbl:models}
\end{table}

\subsection{Prompting}

We present each question to the LLMs in a separate chat session using the template given in Table \ref{tab:prompt-description}.

\begin{table}[!htbp]
    \centering
    \caption{The template used to generate prompts for all questions. The CN text is inserted for CN questions and the multiple answers text is inserted for questions with multiple possible answers.}
    \vspace{0.5em} 
    \begin{tabular}{|p{0.99\textwidth}|}
        \hline
        You are a helpful assistant who answers questions about qualitative
spatial and temporal calculi.

\textless CALCULUS\_DESCRIPTION\textgreater

{[}\textless CONCEPTUAL\_NEIGHBOURHOOD\_DESCRIPTION\textgreater{]}

I will now ask you a question about these relations. {[}There may be
more than one possible relation, in which case name all of the
possible answers.{]} Answer the question and provide the final answer in
the form: "\#\#\# Answer:"

\textless QUESTION\textgreater\\
        \hline
    \end{tabular}
    \label{tab:prompt-description}
\end{table}

The \texttt{CONCEPTUAL\_NEIGHBOURHOOD\_DESCRIPTION}  for regions and intervals is
\emph{``The conceptual neighbourhood of a relation R between entities x and y
is the set of all possible relations which might immediately be
obtained next assuming x and or y were to continuously deform or translate,
without any other relation needing to hold in between.''}

The \texttt{CONCEPTUAL\_NEIGHBOURHOOD\_DESCRIPTION}  for the point-based calcluli PA, STAR and CDC is
\emph{``The conceptual neighbourhood of a relation R between entities x and y
is the set of all possible relations which might immediately be
obtained next assuming x and or y were to continuously translate,
without any other relation needing to hold in between.''}

The 9IM literature refers to topological distance rather than conceptual neighbourhood and so
the \texttt{CONCEPTUAL\_NEIGHBOURHOOD\_DESCRIPTION} for 9IM is
\emph{``The topological distance between two of these topological relationships is the
sum of the absolute values of differences between corresponding
entries of all nine intersections (i.e. the difference is zero if both
entries are empty or both entries are non-empty, and one if they are different). The
closest topological relationships of a relation R between entities x
and y is the set of relations with the least non-zero topological
distances from R.''}

Where an LLM supports setting temperature explicitly, we set temperature to 0.0\footnote{Previous evaluation work on LLMs and QSTR \cite{blackwell2025reproduciblellmevaluationquantifying} has noted better and more predictable performance with zero temperature.} with a fixed
random seed, otherwise we accept the model defaults. Lower temperatures reduce stochasticity in large language model sampling, leading to more deterministic and predictable behaviour at the cost of reduced output diversity. However, it has been shown that no single temperature is optimal for all tasks. Higher temperatures are not always best for creative writing, nor is zero always best for instruction following \citep{li2025exploring}.

\subsection{Repeatability}

Burnell et al.~\cite{burnell2023rethink} call for the routine release of instance-by-instance evaluation results and more granular reporting practices to improve transparency and robustness in AI system evaluations. Accordingly, we publish all our experimental data including all API request / response pairs for all experiments, providing full traceability of all model runs and the opportunity for future re-analysis. The data include the timing of experiments, prompts, metadata, API endpoints, token counts and reasoning traces where available. We provide the scripts used to generate questions and number questions and answers consistently, to allow direct comparison across models.

Models are stochastic in nature and so experimental repeats are essential for statistical comparison. However we must balance costs, and so we follow the guidance given in \citep{blackwell2024towards}, calculating the prediction interval for experimental results where sensible and affordable.

\section{Results}

\begin{figure}
    \centering
    \includegraphics[width=\textwidth]{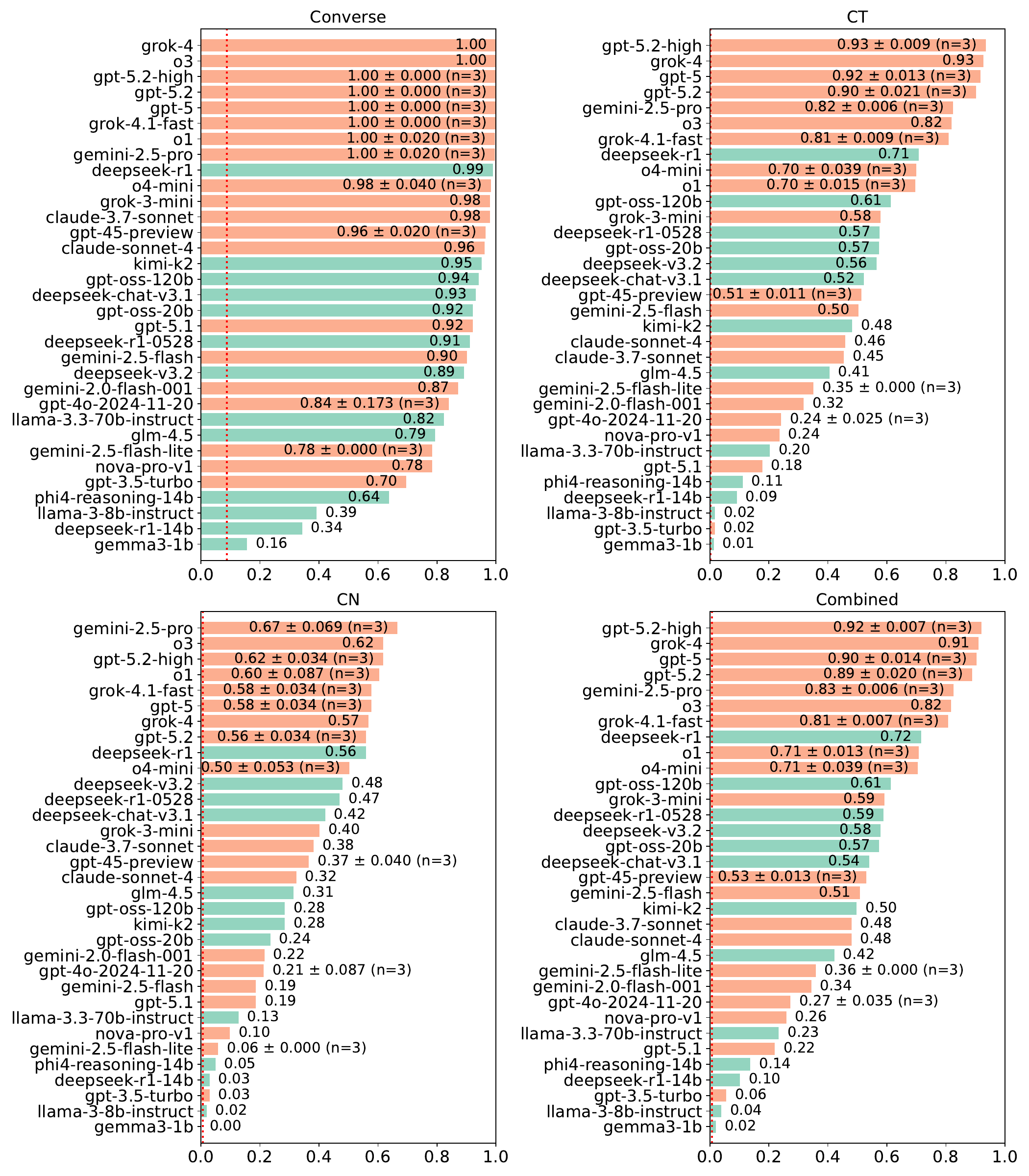}
    \caption{Accuracy of the LLMs tested on our \qstrsmall\ benchmark (converse, CT, CN, and combined questions) using strict evaluation (answers must be precisely correct). The red dotted line is the guess rate. Green bars are open weights models. Experimental repeats are indicated by n=3, but these are not affordable for all models.}
    \label{fig:bake-off}
\end{figure}

In the results tables below, we choose to report micro-averages -- i.e. we simply average the accuracy results across all the selected  individual questions. This has the advantage of simplicity and arguably giving the best overall indication of performance.  We discuss alternatives in  section \ref{sec:limitations} on limitations below.

We  indicate the guess rate in figures with a dashed line.  For questions concerning the converse of a relation, when only a single relation is requested, we compute the guess rate as $1/n$, where $n$ is the number of base relations in the calculus. CN and CT questions both require a set of relations as an answer, so we compute the guess rate as $1/(2^n-1)$ since any non empty subset of base relations could be an answer\footnote{The set of relations in any composition table is always a connected subgraph of the CN, so not all subsets are possible, but we assume an LLM would not know this, so ignore this potential constraint.}. Where the results are a combination of single and multiple answer questions, the guess rate is the micro-average of the individual guess rates.
For the converse experiments, one could take the relation itself as a guess at the converse. For a few calculi this would result in a higher accuracy for the guess; this would be the case in RCC-5, RCC-8, RCC-22, and 9IM; these calculi have a number of relations -- all those which are not proper part relationships (or P-INSIDE in the case of RCC-22) --  which are their own converse (2/5, 4/8, 6/22, 4/8 relations respectively) and  these values are all larger than $1/n$.  For all the other  calculi, there would be no difference in the computed guess rate, since only equality is its own converse. Another slightly higher baseline guess rate would be to only choose guesses which form a connected subgraph of the CN since  CTs always appear to obey this constraint \citep{cohn1994comparison}. Note that the model results for converse are all sufficiently high that even for the best of these alternative converse guess rates, they still substantially outperform it -- see Fig. \ref{fig:accuracy-by-calculus}.   Other guess rate baselines for the CT and CN cases are possible -- e.g. only considering subsets of a particular  cardinality, but there seems no obvious theoretical justification for these.

All models tested achieve higher accuracy than the guess rate for the combined Converse, CT and CN benchmark (Fig. \ref{fig:bake-off}).
\bestmodel\ with high reasoning \repeats{1659}{1666}{1663}{1806}\footnote{Where we had the resources to do three experimental repeats, we use the notation (a:b:c/d) to mean a/d, b/d and c/d.} is the best performing model.
Of the top ten models, only one (\emph{DeepSeek R1}) is an open weights model.


It is notable that \emph{GPT-5.1} performs poorly (0.22) but both the preceding GPT version (\emph{GPT-5}) and the subsequent GPT version (\gptfivetwo)
are amongst the best performers (0.90 and 0.89 respectively).
\emph{GPT-5.1} was the first OpenAI release to use ``adaptive reasoning'' to decide how much to think before responding to questions\footnote{OpenAI Product Release \url{https://openai.com/index/gpt-5-1/}, retrieved April 2026.}. In our tests,
\emph{GPT-5.1} used a mean of only 24 completion tokens per benchmark question (so was clearly not using chain of thought reasoning),
whereas \emph{GPT-5} used a mean of 4149 tokens (172x) and \gptfivetwo\ 1845 tokens (77x). Our results show that \gptfivetwo\ with high
reasoning effort outperforms all our other GPT model experiments, so perhaps OpenAI were still refining the question complexity classification algorithm in the \emph{GPT-5.1} release.


\begin{figure}
    \centering
    \includegraphics[width=\textwidth]{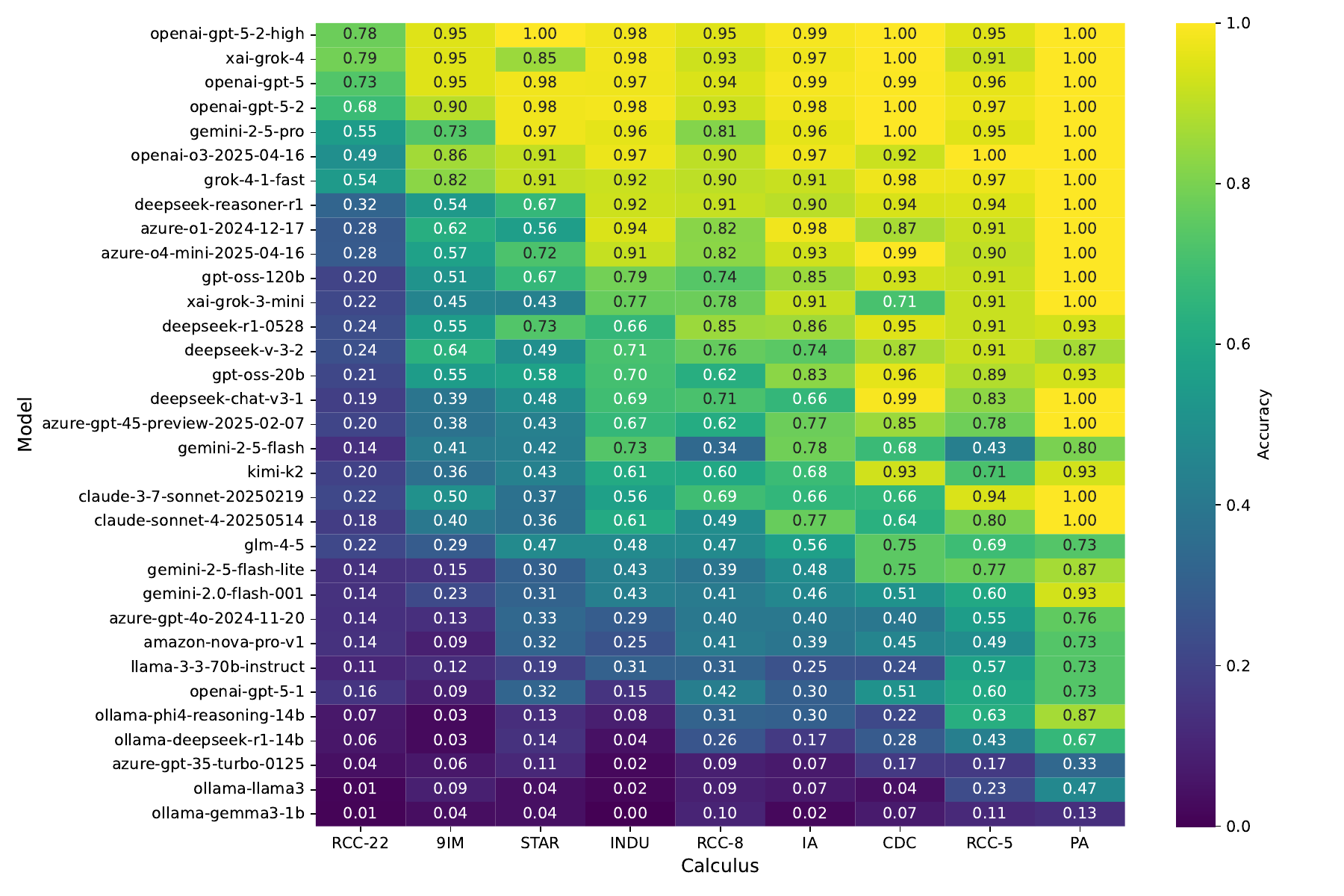}
    \caption{Accuracy by model by calculus. }
    \label{fig:accuracy-by-model-by-calculus}
\end{figure}

Although setting high reasoning effort in \gptfivetwo\ improves performance
across most calculi, it makes RCC-5 performance slightly worse (0.96
to 0.95, Fig. \ref{fig:accuracy-by-model-by-calculus}) and reduces the percentage of correct and consistent RCC-5 answers
across all three repeats from 97\% to 91\%.
\gptfivetwo\ with high reasoning effort gives consistent and correct answers
across all three repeats of CDC, PA and STAR questions but answers only 67\%
of RCC-22 questions correctly and consistently across all three repeats
(despite total RCC-22 accuracy being 78\%).  This inconsistency might indicate guessing, or at least a
lack of confidence in the RCC-22 answers given.

We tested two small models, on local hardware, using Ollama\footnote{\url{https://ollama.com}, accessed March 2026} (\emph{Gemma 3:1B} and \emph{Llama 3:8B}), and these were the two worst performing models.

There has been an extraordinary improvement in model accuracy over time, e.g. OpenAI models improved from 0.06 \gptthreefive\ to 0.92 (\gptfivetwo) in two years (Fig. \ref{fig:bake-off}, Table \ref{tbl:models}).

Models typically perform best on converse questions, followed by CT questions, with CN questions being the most challenging (Fig. \ref{fig:bake-off}).

\begin{figure}
    \centering
    \includegraphics[width=\textwidth]{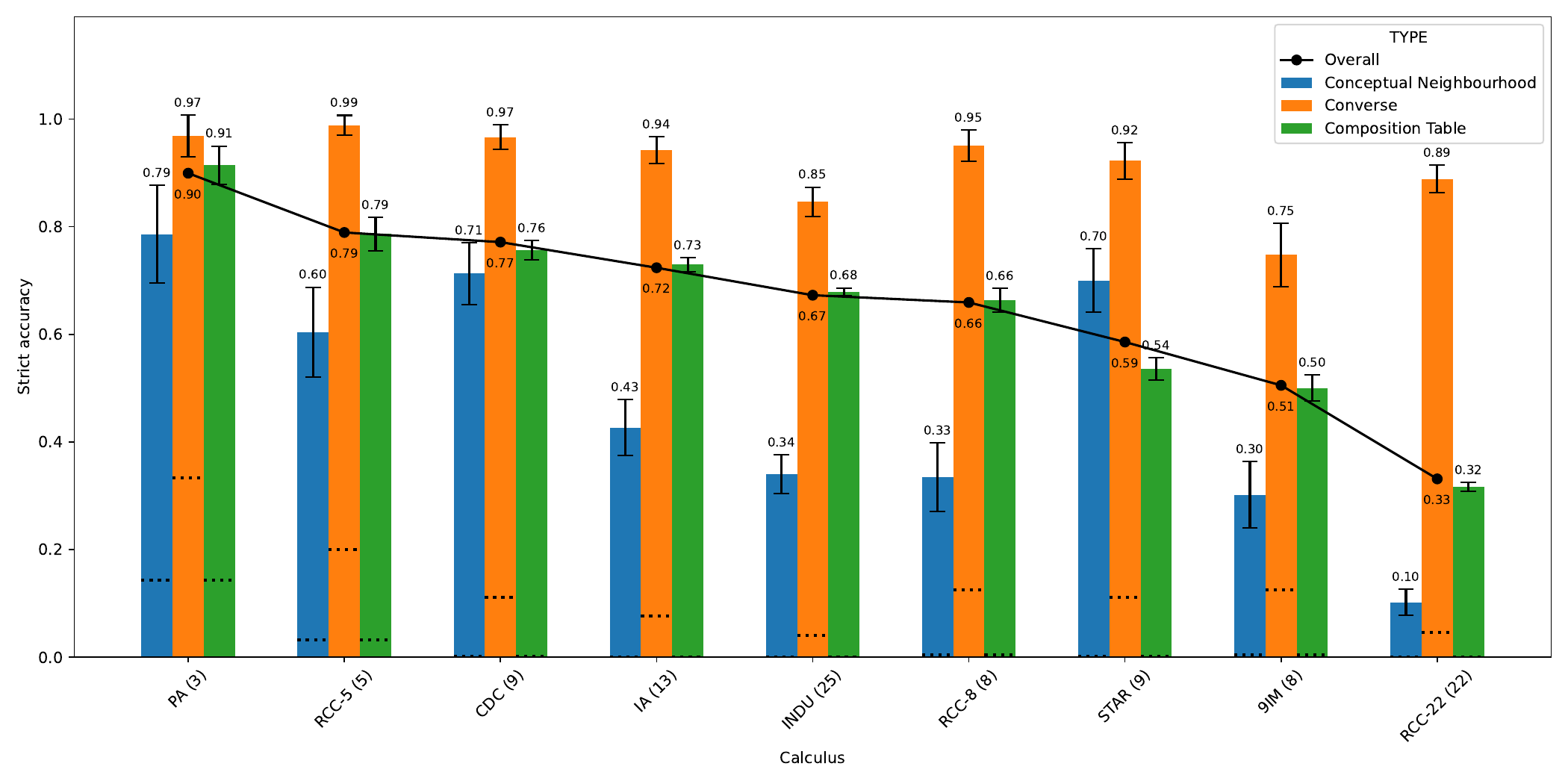}
    \caption{Accuracy by calculus (solid line), broken down by Composition Table, Conceptual Neighbourhood and Converse questions (bars), for all LLMs tested. Dotted lines
    show the guess rate. Error bars show the prediction interval. The number of base relations for a calculus is shown in brackets beside the label.}
    \label{fig:accuracy-by-calculus}
\end{figure}

Overall, LLMs were most accurate with the PA calculus, and least accurate with RCC-22 (Fig. \ref{fig:accuracy-by-calculus}).
For all calculi except STAR, the LLMs were more accurate with CT than CN (but the CN for STAR is perhaps obvious because of the numeric ordering of relations).

For RCC, accuracy declines with the number of base relations (RCC-5 is more accurate than RCC-8, which is more accurate than RCC-22) but in general, accuracy is not proportional to the number of base relations (Fig. \ref{fig:accuracy-by-calculus}). For example, INDU has higher accuracy than RCC-22, despite it having three more base relations and having the highest number
of base relations of all calculi tested. Also, STAR has a higher accuracy than 9IM (9 vs 8 relations) and CDC and IA have a higher accuracy than RCC-8 and 9IM (9 and 13 relations respectively compared to 8 and 8).

\begin{table}[!htbp]
\caption{Number of False Positive (relations specified incorrectly) and False Negative (relations not specified that should have been specified) in CN and CT answers by model. (Where there are multiple repeats, numbers are divided by the number of repeats for model-by-model direct comparison). TP = True Positive, FP = False Positive, and FN = False Negative. The higher of FP or FN is shown in bold.}
\label{tab:fpfn}
\vspace{0.5em} 
\resizebox{\textwidth}{!}{%
\begin{tabular}{lllllll}
\toprule
model & CN TP & CN FP & CN FN & CT TP & CT FP & CT FN \\
\midrule
openai-gpt-5-2-high & 303.7 & 36.3 & \textbf{50.3} & 7706.0 & \textbf{245.3} & 138.0 \\
x-ai/grok-4 & 306 & \textbf{57} & 48 & 7513 & 76 & \textbf{331} \\
openai/gpt-5 & 298.3 & 39.0 & \textbf{55.7} & 7659.3 & \textbf{239.0} & 184.7 \\
openai/gpt-5.2 & 298.3 & 36.0 & \textbf{55.7} & 7692.0 & \textbf{492.7} & 152.0 \\
google/gemini-2.5-pro & 305.7 & 34.0 & \textbf{48.3} & 6969.0 & \textbf{935.3} & 875.0 \\
openai/o3 & 301 & 47 & \textbf{53} & 7256 & \textbf{701} & 588 \\
x-ai/grok-4.1-fast & 294.7 & 39.0 & \textbf{59.3} & 6130.3 & 269.0 & \textbf{1713.7} \\
deepseek/deepseek-r1 & 299 & \textbf{72} & 55 & 5160 & 396 & \textbf{2684} \\
openai/o1 & 289.7 & 31.0 & \textbf{64.3} & 5626.7 & 954.7 & \textbf{2217.3} \\
openai/o4-mini & 267.7 & 43.3 & \textbf{86.3} & 5602.3 & 980.7 & \textbf{2241.7} \\
openai/gpt-oss-120b & 250 & \textbf{114} & 104 & 4846 & 1384 & \textbf{2998} \\
x-ai/grok-3-mini & 272 & 75 & \textbf{82} & 3843 & 669 & \textbf{4001} \\
deepseek/deepseek-r1-0528 & 288 & \textbf{68} & 66 & 4342 & 441 & \textbf{3502} \\
deepseek/deepseek-v3.2 & 287 & \textbf{79} & 67 & 5184 & 1276 & \textbf{2660} \\
openai/gpt-oss-20b & 261 & \textbf{219} & 93 & 4712 & 1016 & \textbf{3132} \\
deepseek/deepseek-chat-v3.1 & 263 & 57 & \textbf{91} & 3903 & 907 & \textbf{3941} \\
openai/gpt-45-preview & 262.3 & \textbf{96.3} & 91.7 & 4818.0 & 1128.0 & \textbf{3026.0} \\
google/gemini-2.5-flash & 260 & \textbf{206} & 94 & 4997 & 2682 & \textbf{2847} \\
moonshotai/kimi-k2 & 238 & 94 & \textbf{116} & 3295 & 967 & \textbf{4549} \\
anthropic/claude-3.7-sonnet & 270 & \textbf{89} & 84 & 5407 & 2085 & \textbf{2437} \\
anthropic/claude-sonnet-4 & 269 & \textbf{91} & 85 & 4675 & 1545 & \textbf{3169} \\
z-ai/glm-4.5 & 249 & \textbf{113} & 105 & 3654 & 1309 & \textbf{4190} \\
google/gemini-2.5-flash-lite & 184.0 & \textbf{254.0} & 170.0 & 2639.0 & 1656.0 & \textbf{5205.0} \\
google/gemini-2.0-flash-001 & 257 & \textbf{156} & 97 & 4235 & 2738 & \textbf{3609} \\
openai/gpt-4o-2024-11-20 & 219.7 & 105.0 & \textbf{134.3} & 1733.0 & 1061.0 & \textbf{6111.0} \\
amazon/nova-pro-v1 & 216 & \textbf{221} & 138 & 1830 & 1157 & \textbf{6014} \\
meta-llama/llama-3.3-70b-instruct & 251 & \textbf{208} & 103 & 2585 & 1499 & \textbf{5259} \\
openai/gpt-5.1 & 249 & \textbf{178} & 105 & 2325 & 3246 & \textbf{5519} \\
microsoft/phi4-reasoning-14b & 44 & 160 & \textbf{310} & 256 & 2620 & \textbf{7588} \\
deepseek/deepseek-r1-distill-qwen-14b & 41 & 248 & \textbf{313} & 189 & 3565 & \textbf{7655} \\
openai/gpt-3.5-turbo & 137 & 171 & \textbf{217} & 1507 & 2825 & \textbf{6337} \\
meta-llama/llama-3-8b-instruct & 120 & 208 & \textbf{234} & 1457 & 3231 & \textbf{6387} \\
google/gemma3-1b & 106 & \textbf{569} & 248 & 958 & 4304 & \textbf{6886} \\
\bottomrule
\end{tabular}%
}
\end{table}

For most models (28/33) there are more false negative answers than false positives to CT questions, meaning
that those models under-specify relations in CT answers rather than provide too many relations (Table \ref{tab:fpfn}); it is notable that for the best six models overall, this pattern is reversed, with only \emph{Grok 4} having a higher FN than FP count -- thus these models are much more likely to over predict than under predict relations to include in a CT entry.
However this is not the pattern for CN questions where results are mixed, though again the top six models all under predict, again with the exception of  \emph{Grok 4}.


The converse of an identity relation is always the identity relation, and so these questions should be straightforward for LLMs. However our four worst performing models (\gptthreefive, \emph{DeepSeek R1 14b}, \emph{Gemma3 1b}, and \emph{llama3}) get some answers incorrect.


In a CT, the composition of any relation \emph{R} with the identity relation is the relation \emph{R} unchanged, and so these questions should also be straightforward for LLMs. However, only five of the models tested (\emph{Claude 3.7 Sonnet}, \emph{GPT-5-2} (with both default and high reasoning), \emph{o3} and \emph{Grok 4}) get all such answers correct.

We now look at the results calculus by calculus.

\subsection{PA (3 relations, 15 questions)}
The Point Algebra (PA) is a qualitative temporal calculus for representing and reasoning about temporal relationships between points on the real number line.
The theoretical basis and computational properties of PA were formalised by \citet{vilain1986constraint}, who introduced it as part of a general framework for temporal constraint satisfaction.
It consists of three jointly exhaustive and pairwise disjoint binary spatial relations. \textless{} means that x is strictly earlier than y. = means that x and y are coincident. \textgreater{} means that x is strictly later than y.
As such, it is has the fewest base relations (three) of all the calculi considered here and is the most accurate across all LLMs tested (Fig. \ref{fig:accuracy-by-calculus}).

About half of the models tested (16/33), including \gptfivetwo, get all PA questions correct (Fig. \ref{fig:accuracy-by-model-by-calculus}).
\gptfivetwo\ with high reasoning effort uses a mean and standard deviation of $31 \pm 22$ reasoning tokens to solve PA questions involving
\emph{=}, $66 \pm 55$ for questions involving \emph{>} and $67 \pm 61$ for questions involving \emph{<}. Although the number of reasoning tokens used for each question is highly variable, this result is consistent with \emph{=}
perhaps being intuitively easier than the other two relations.


One might expect the CT question \emph{If =(x,y) and =(y,z) then what are the possible relations between x and z?}
to be straightforward (the answer is only \emph{=}), but \gptthreefive\ (10/15) answered with all relations \emph{\textless{}, =, and \textgreater{}}, suggesting that it has no real intuition about PA.
Similarly, \kimi\ (14/15) answers the question \emph{If \textgreater{}(x,y) and \textgreater{}(y,z) then what are the possible relations between x and z?} incorrectly
with all relations \emph{\textless{}(x,z);=(x,z);\textgreater{}(x,z)}, rather than \emph{\textgreater{}(x,z)}, showing a lack of understanding about transitivity.

\subsection{RCC-5 (5 relations, 35 questions)}

\begin{figure}
    \centering
    \includegraphics[width=0.5\textwidth]{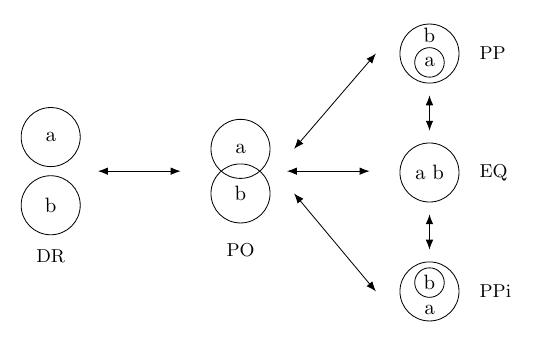}
    \caption{The five relations of the RCC-5 calculus, depicted with its conceptual neighbourhood: \DR\ (Discrete), \PO\ (Partially Overlapping), \PP\ (Proper Part) and \EQ\ (Equals). \PPi\ is the converse of \PP. }
    \label{fig:rcc5}
\end{figure}

The Region Connection Calculus, RCC-5 is a qualitative spatial calculus for representing and reasoning about spatial relationships between\footnote{There are other interpretations of RCC-5 involving other kinds of spatial entities, but here we stick with this model; similar comments apply to some of the other calculi below.} non-empty regions in a topological space (\cite{bennett1994spatial, jonsson1997}, Fig. \ref{fig:rcc5}).

Despite having just five base relations, only one model (\emph{o3}) got all RCC-5 answers correct (Fig. \ref{fig:accuracy-by-model-by-calculus}). The majority of models (31/33) got all RCC-5 converse answers completely correct, but only four (\emph{Gemini 2.5 pro}, \emph{DeepSeek R1}, \emph{o3}, and \emph{Deepseek V3.2}) got the all CN questions correct and only three (\emph{o3}, \gptfivetwo, and \emph{GPT 5} - all OpenAI Models) all CT questions.

\gptfivetwo\ with high reasoning effort always failed to correctly answer the CN question for PO -- always failing
to specify \EQ\ but correctly specifying \DR, \PP, and \PPI. We see a similar pattern for RCC-8, later.

\gptfivetwo\ with high reasoning effort uses a mean and a standard deviation of $496 \pm 434$ reasoning tokens to solve RCC-5 questions involving \PPI,
$455 \pm 306$ for \PO,
$401 \pm 366$ for \DR,
$304 \pm 209$ for \PP,
and only $96 \pm 70$ for \EQ.
It is unsurprising
that \EQ\ needs less reasoning effort, but why \PPI\ needs the most reasoning effort remains unclear and in particular why it is much harder than \PP. We see a similar pattern between \TPPi\ and \TPP\ as well as \NTPPi\ and \NTPP\ in RCC-8.


One might expect the CT question \emph{If EQ(x,y) and EQ(y,z) then what are the possible relations between x and z?} to be straightforward (the answer is only \emph{EQ}) but \emph{Gemma 3 1b} answers with \emph{the possible relations between x and z are: EQ(x,y); EQ(y,z); PP(x,y); PP(y,z); PO(x,y); PO(y,z); DR(x,y); DR(y,z)}, showing
a complete lack of understanding of the calculus or composition.
Similarly, for the CT question \emph{If PP(x,y) and PP(y,z) then what are the possible relations between x and z?} (answer is only \emph{PP}),
four models get this wrong -- \emph{GPT4o} (\emph{PP, EQ}), \emph{Gemini 2.5 Flash} (\emph{PP, DR, PO, EQ}),
\gptthreefive\ (\emph{DR, PO, EQ}) and \emph{Gemma 3b} (\emph{PP, PP, EQ}).

\subsection{CDC (9 relations, 99 questions)}

\begin{figure}
    \centering
    \includegraphics[width=0.5\textwidth]{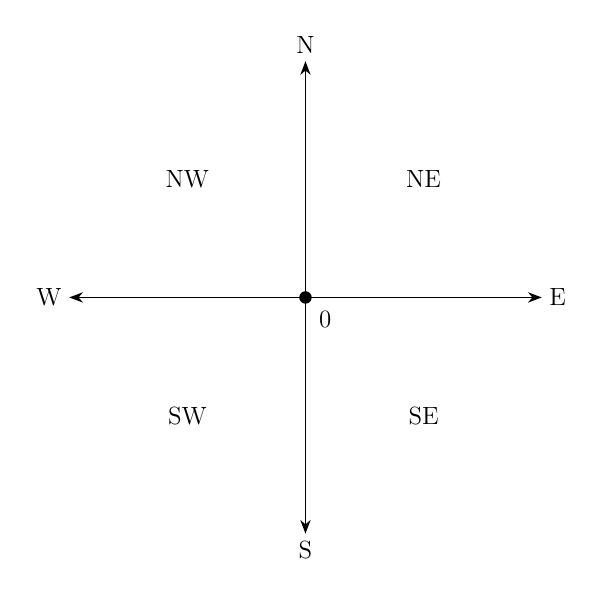}
    \caption{Depiction of the nine relations of the Cardinal Direction Calculus (CDC). For example N(x,y) means that x lies along the line that extends due north of y. NE(x,y) means that x lies to the east of the line that extends due north of y, and to the north the line that extends due east of y, and so on. }
    \label{fig:cdc}
\end{figure}

The Cardinal Direction Calculus (CDC) is a qualitative spatial reasoning calculus for representing and reasoning about the direction relationships between pairs of points in the Euclidean plane.
Unfortunately, there are several calculi referred to in the literature as Cardinal Direction Calculus (CDC), each with nuanced semantics,
including \cite{frank1996qualitative} and \cite{ligozat1998reasoning}.
The CDC we use here is based on Frank's projection-based directions . However, Frank's CT (Table 4 in \cite{frank1996qualitative}) has a typo -- N followed by NE should be NE rather than N (with this change his composition table becomes symmetrical about the leading diagonal). We expand \emph{e} to \emph{NE E SE}, \emph{s} to \emph{SW S SE}, \emph{w} to \emph{NW W SW} and \emph{n} to \emph{NW N NE}. We expand \emph{o} depending on context; \emph{S} followed by \emph{N, o} becomes \emph{0 S N}, and for \emph{NW} followed by \emph{SE}, \emph{o} becomes \emph{0 S N W E SW NW SE NE} etc. The resulting calculus is depicted in Fig. \ref{fig:cdc}.


Only three of the models (\gptfivetwo\ (both with default and high reasoning), \emph{Grok-4}, and \emph{Gemini-2.5-Pro} get all CDC questions correct (Fig. \ref{fig:accuracy-by-model-by-calculus}), but most models (30/33) get all the CDC converse questions correct. We suspect that opposite compass directions occur
frequently in the training data.

One might expect the CT question \emph{If N(x,y) and N(y,z) then what are the possible relations between x and z?} to be straightforward (the answer is only \emph{N}) but
four models (\emph{ollama-llama3}, \emph{GPT-4o}, \gptthreefive\, and \emph{deepseek-r1:14b}) get this incorrect. Mistakes are similarly seen for \emph{S}, \emph{E} and \emph{W}.

\begin{figure}
    \centering
    \includegraphics[width=0.49\textwidth]{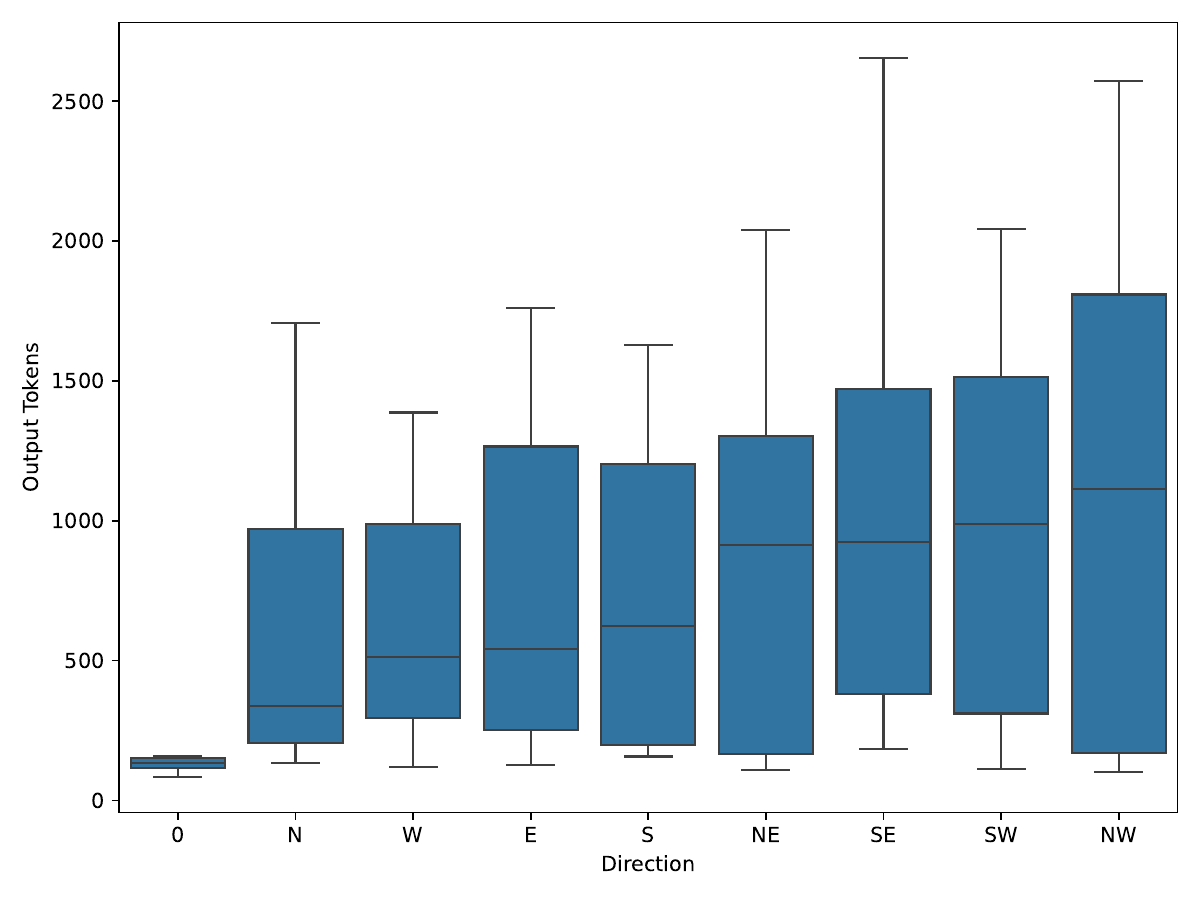}\hfill
    \includegraphics[width=0.49\textwidth]{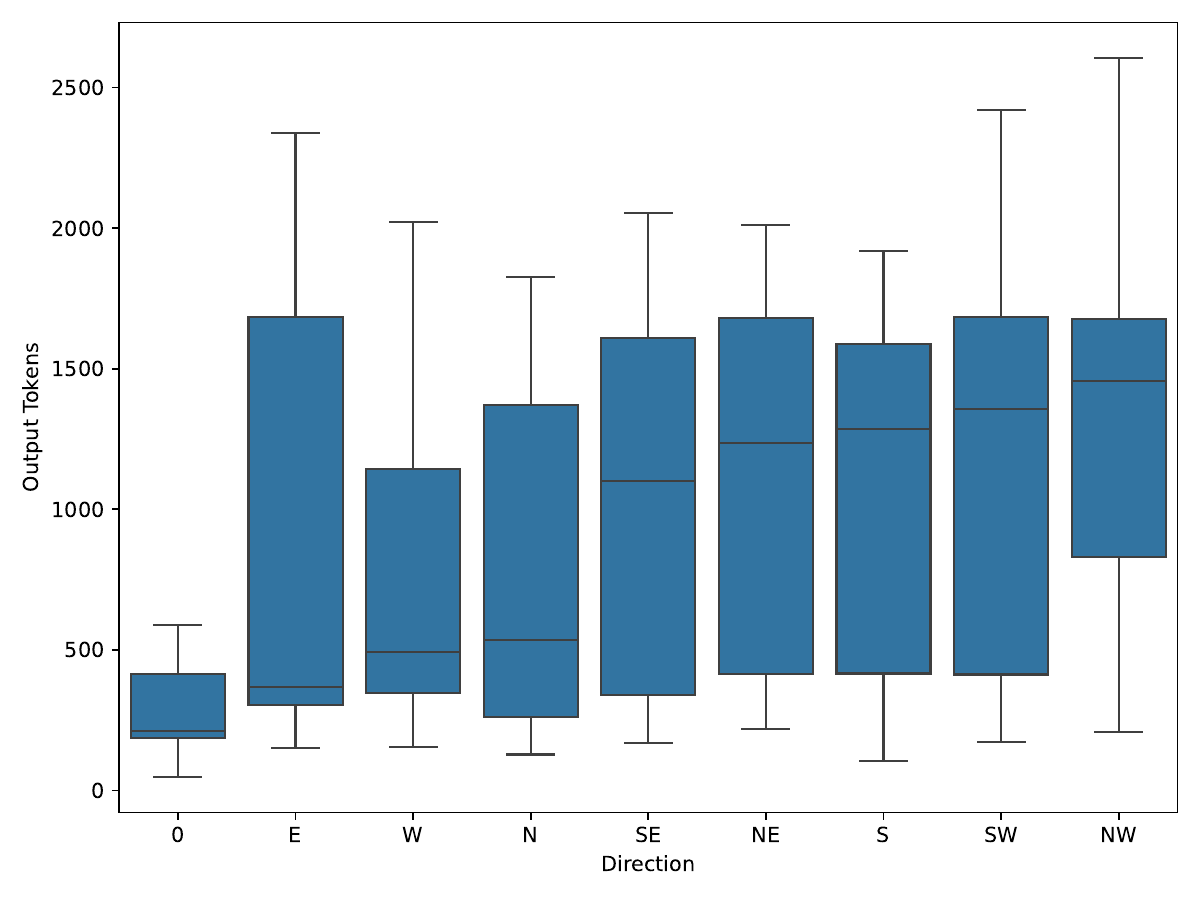}
    \caption{Distribution of output tokens by correct answer direction for kimi-k2 on CDC - eponymous relations on the left, nonce word relations on the right. Note that intercardinal directions use more tokens than cardinal directions in the eponymous case. The effect is somewhat true for nonce relations on the right except that South is an anomaly.}
    \label{fig:cdc-tokens}
\end{figure}

For CDC, \relation{0} is the most accurate relation, and as in our earlier work \citep{cohn2025evaluatingabilitylargelanguage}, inter-cardinal directions tend to require more output tokens than cardinal directions (for example, see Fig. \ref{fig:cdc-tokens}). We suggest that problems involving intercardinal directions can be thought of as a composition of cardinal directions and are thus two-hop reasoning problems rather than one-hop\footnote{There are a few natural languages, e.g. Finnic languages, Sanskrit and Breton, where the intercardinals do not have names composed of pure cardinal direction names -- it would be interesting to see if this pattern repeats -- but these are all low resource languages thus creating issues in testing this hypothesis.}.

\subsection{IA (13 relations, 195 questions)}

\begin{table}[!htbp]
\centering
\caption{The thirteen relations from Allen's Interval Algebra \citep{allen1983maintaining}.}
\vspace{0.5em} 
\renewcommand{\arraystretch}{1.4} 
\begin{tabular}{>{\raggedright}p{3cm}ccl}
\hline
\textbf{Relation} & \textbf{Symbol} & \textbf{Symbol for Converse} & \textbf{Schematic} \\
\hline
$X$ before $Y$    & $<$  & $>$  & XXX \hspace{0.5cm} YYY \\
$X$ equal $Y$     & $=$  & $=$  & \begin{tabular}{@{}l@{}}XXX \\ YYY\end{tabular} \\
$X$ meets $Y$     & m    & mi   & XXXYYY \\
$X$ overlaps $Y$  & o    & oi   & \begin{tabular}{@{}l@{}}XXX \\ \hspace{0.2cm}YYY\end{tabular} \\
$X$ during $Y$    & d    & di   & \begin{tabular}{@{}l@{}}~~~XXX \\ YYYYYYY\end{tabular} \\
$X$ starts $Y$    & s    & si   & \begin{tabular}{@{}l@{}}XXX \\ YYYYYY\end{tabular} \\
$X$ finishes $Y$  & f    & fi   & \begin{tabular}{@{}l@{}}~~~~~~XXX \\ YYYYYY\end{tabular} \\
\hline
\end{tabular}
\label{tab:allen-relations}
\end{table}

Allen's Interval Algebra (IA, \cite{allen1983maintaining}) is a qualitative temporal calculus for representing and reasoning about one-piece closed non-empty temporal intervals on the real number line. It consists of thirteen jointly exhaustive and pairwise disjoint binary temporal relations, between an interval $x$ and an interval $y$ (Table \ref{tab:allen-relations}).

None of the LLMs tested gets all IA questions correct (Fig. \ref{fig:accuracy-by-model-by-calculus}), but many models (22/33) get all the IA converse questions correct .

\gptfivetwo\ with high reasoning is the best performing model  \repeats{193}{193}{192}{195}. Of the seven incorrect answers, six are CN and one is CT.
Two CN questions have repeated errors: \emph{What are the conceptual neighbours of d(x,y)?} and \emph{What are the conceptual neighbours of =(x,y)?} -- both give the correct relations but provide additional, incorrect relations.

\subsection{RCC-8 (8 relations 80 questions)}

The Region Connection Calculus, RCC-8 is a qualitative spatial calculus for representing and reasoning about spatial relationships between non-empty regular closed regions of uniform dimension in a topological space (\cite{randell1992spatial,cohn1997qualitative}, Fig. \ref{rcc8-cn-diagram}).


None of the LLMs tested gets all RCC-8 questions correct (Fig. \ref{fig:accuracy-by-model-by-calculus}), but many models (25/33) get all the RCC-8 converse questions correct.
\emph{GPT-5} is the best performing model \repeats{74}{75}{77}{80}. Of the 14 incorrect answers, 13 were CN and one CT. The incorrect CT answer
\emph{If TPPi(x,y) and NTPP(y,z) then what are the possible relations between x and z?} was missing the \emph{TPPi} relation. In 8/13 incorrect CN answers, all the relations were correctly listed, but there was also incorrect additional relations in the
answer. In 9/13 the error included the incorrect presence or absence of \emph{EQ}.

\gptfivetwo\ with high reasoning effort always failed to correctly answer the CN question for PO -- failing
to specify \EQ\ but correctly specifying \EC, \TPP, and \TPPi. RCC-5 showed a similar pattern, above.

\gptfivetwo\ with high reasoning effort uses a mean and a standard deviation of $1267 \pm 909$ reasoning tokens to solve RCC-8 questions involving \TPPi, $1053 \pm 896$ involving \TPP, but only $686 \pm 623$ for \DC\ and $132 \pm 174$ for \EQ.
\TPPi\ is perhaps intuitively more difficult than the other RCC-8 relations.
As previously noted for RCC-5 with \DR, the fewer reasoning tokens for \DC\ aligns with the results reported by Ragni et al \cite{ragni2007cross} that humans seem to prefer mental models involving \DC.

Some models (10/33) sometime fail to follow the rubric or answer with invalid relation names, e.g. o4-mini answers \emph{TPPPi} (note three Ps not two!) once
and \emph{NTPPPi} once, kimi-k2 answers \emph{TTPi} once, suggesting issues with robustly reasoning with tokenised symbols of this form.

\sloppypar Apart   from \emph{LaTeX description Prefix symbols} (accuracy 0.65),  \emph{Schematic description Prefix nonce words} (0.53), and to a lesser extent \emph{Textual Description Infix Words}(accuracy 0.71) the description style makes little difference to o1 accuracy with the RCC-8 questions (Fig. \ref{fig:rcc-8-extras-o1}). Even when no description of the calculus is given, accuracy is still 0.81, suggesting that o1 is
able to generalise and draw upon both its training data and the information given in the prompt. When the RCC-8 relation names are swapped, accuracy is still 0.80 suggesting that o1 can use information given in the prompt to override training data, either by explicitly reasoning, or by exploiting semantic similarities between the descriptions of the relations in the prompt and those in the literature.
LaTeX markup perhaps obscures the underlying
concepts and is almost certainly less prevalent in the training data\footnote{Although the pdf of papers generated with latex descriptions of the RCC-8 definitions are in the public domain, we do not believe the latex sources are.}, perhaps accounting for its poorer performance.
In the case of \emph{Schematic description Prefix nonce words}, RCC-8 is never mentioned and the RCC-8 relations are not explicitly described, nor are we aware of papers about RCC-8 using this description style, so the model has little recourse to training data, suggesting that o1 is less able to reason with schematic diagrams.

\begin{figure}
    \centering
    \includegraphics[width=0.8\textwidth]{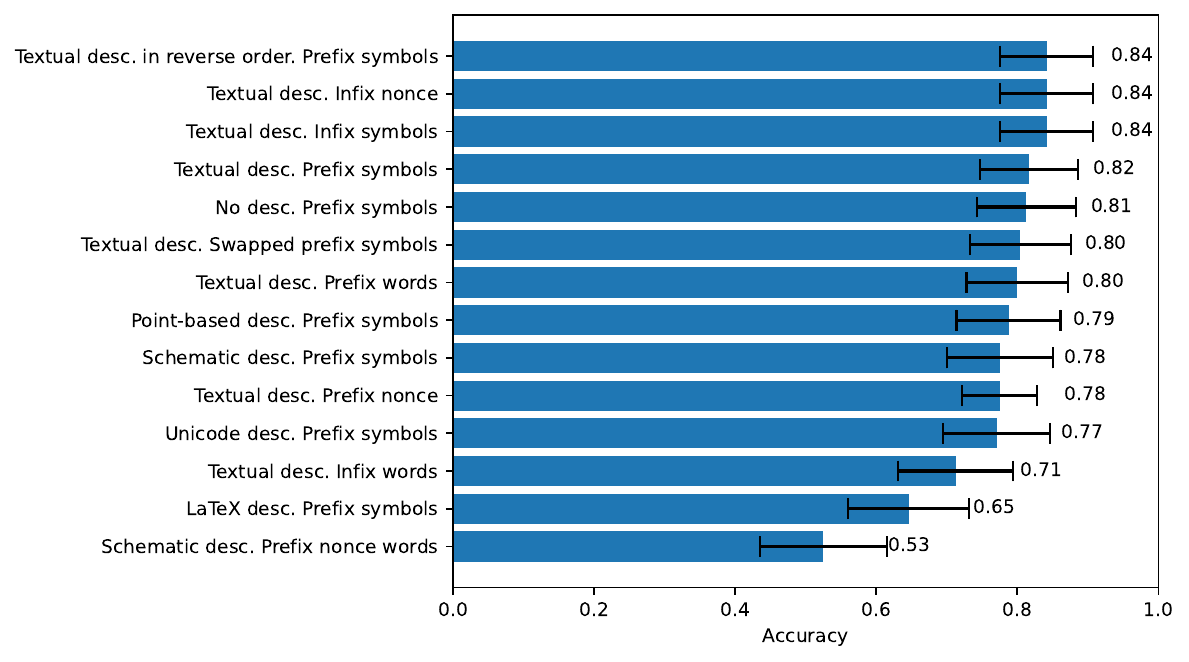}
    \caption{Accuracy of RCC-8 answers by description style for o1 (3 repeats of n=80 questions per bar). The error bar is the prediction interval.}
    \label{fig:rcc-8-extras-o1}
\end{figure}

\subsection{INDU (25 relations, 675 questions)}

\begin{table}[!htbp]
\centering
\begingroup
\newcommand{\xb}{X^{\mathrm{b}}}
\newcommand{\xe}{X^{\mathrm{e}}}
\newcommand{\xd}{X^{\mathrm{d}}}
\newcommand{\yb}{Y^{\mathrm{b}}}
\newcommand{\ye}{Y^{\mathrm{e}}}
\newcommand{\yd}{Y^{\mathrm{d}}}

\caption{The 25 INDU relations from \cite{pujari1999indu} with corrections. The $b$ and $e$ superscripts represent the beginning and end points of the respective interval and a $d$ superscript its duration. }
\vspace{0.5em} 
\label{tab:indu}
\resizebox{\textwidth}{!}{%
\begin{tabular}{ccll}
\toprule
\textbf{Relation} & \textbf{Converse} & \textbf{Schematic} & \textbf{End-point relations \& relative duration} \\
\midrule
$b^{<}$ & $bi^{>}$ &
\makecell[l]{\texttt{XXX}\\[-0.3ex]\texttt{~~~~~YYYY}} &
$\xb<\yb,\ \xb<\ye,\ \xe<\yb,\ \xe<\ye,\ \xd<\yd$ \\

$b^{=}$ & $bi^{=}$ &
\makecell[l]{\texttt{XXX}\\[-0.3ex]\texttt{~~~~~YYY}} &
$\xb<\yb,\ \xb<\ye,\ \xe<\yb,\ \xe<\ye,\ \xd=\yd$ \\

$b^{>}$ & $bi^{<}$ &
\makecell[l]{\texttt{XXXX}\\[-0.3ex]\texttt{~~~~~YYY}} &
$\xb<\yb,\ \xb<\ye,\ \xe<\yb,\ \xe<\ye,\ \xd>\yd$ \\

$m^{<}$ & $mi^{>}$ &
\makecell[l]{\texttt{XXX}\\[-0.3ex]\texttt{~~~YYYY}} &
$\xb<\yb,\ \xb<\ye,\ \xe=\yb,\ \xe<\ye,\ \xd<\yd$ \\

$m^{=}$ & $mi^{=}$ &
\makecell[l]{\texttt{XXX}\\[-0.3ex]\texttt{~~~YYY}} &
$\xb<\yb,\ \xb<\ye,\ \xe=\yb,\ \xe<\ye,\ \xd=\yd$ \\

$m^{>}$ & $mi^{<}$ &
\makecell[l]{\texttt{XXXX}\\[-0.3ex]\texttt{~~~~YYY}} &
$\xb<\yb,\ \xb<\ye,\ \xe=\yb,\ \xe<\ye,\ \xd>\yd$ \\

$o^{<}$ & $oi^{>}$ &
\makecell[l]{\texttt{XXX}\\[-0.3ex]\texttt{~YYYY}} &
$\xb<\yb,\ \xb<\ye,\ \xe>\yb,\ \xe<\ye,\ \xd<\yd$ \\

$o^{=}$ & $oi^{=}$ &
\makecell[l]{\texttt{XXX}\\[-0.3ex]\texttt{~YYY}} &
$\xb<\yb,\ \xb<\ye,\ \xe>\yb,\ \xe<\ye,\ \xd=\yd$ \\

$o^{>}$ & $oi^{<}$ &
\makecell[l]{\texttt{XXXX}\\[-0.3ex]\texttt{~~YYY}} &
$\xb<\yb,\ \xb<\ye,\ \xe>\yb,\ \xe<\ye,\ \xd>\yd$ \\

$d^{<}$ & $di^{>}$ &
\makecell[l]{\texttt{~~XXX~~}\\[-0.3ex]\texttt{YYYYYY}} &
$\xb>\yb,\ \xb<\ye,\ \xe>\yb,\ \xe<\ye,\ \xd<\yd$ \\

$s^{<}$ & $si^{>}$ &
\makecell[l]{\texttt{XXX~~}\\[-0.3ex]\texttt{YYYYY}} &
$\xb=\yb,\ \xb<\ye,\ \xe>\yb,\ \xe<\ye,\ \xd<\yd$ \\

$f^{<}$ & $fi^{>}$ &
\makecell[l]{\texttt{~~XXX}\\[-0.3ex]\texttt{YYYYY}} &
$\xb>\yb,\ \xb<\ye,\ \xe>\yb,\ \xe=\ye,\ \xd<\yd$ \\

$eq^{=}$ & - &
\makecell[l]{\texttt{XXXX}\\[-0.3ex]\texttt{YYYY}} &
$\xb=\yb,\ \xb<\ye,\ \xe>\yb,\ \xe=\ye,\ \xd=\yd$ \\
\bottomrule
\end{tabular}%
}
\endgroup
\end{table}

INDU is a qualitative temporal calculus for representing and reasoning about one-piece closed non-empty temporal intervals on the real number line (\cite{pujari1999indu}, Table \ref{tab:indu}); it is a refinement of the IA, where the relations which do not imply a part/equals relation are split into three cases, distinguishing where one interval is larger, equal to, or smaller than the second one.


\begin{figure}
    \centering
    \includegraphics[width=0.99\textwidth]{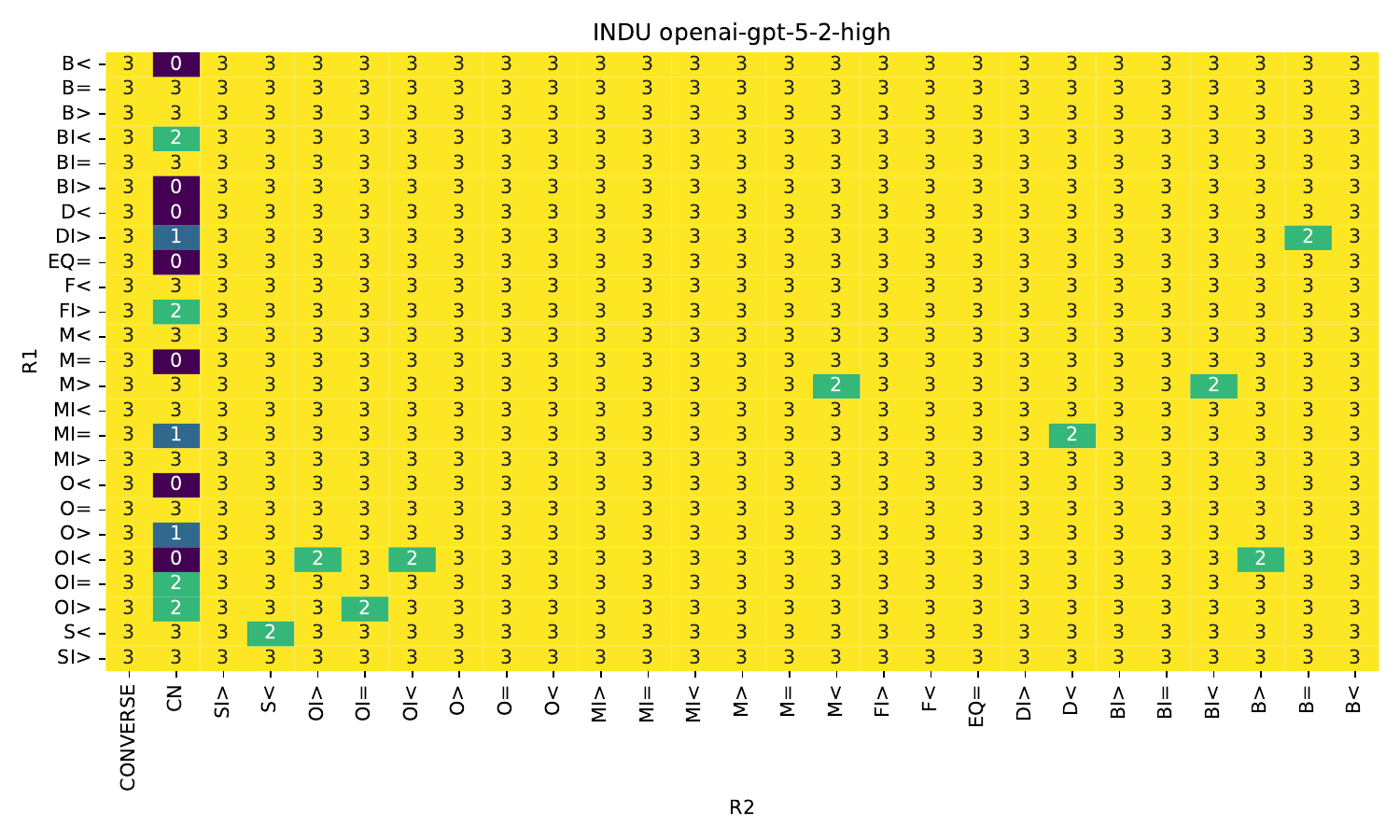}
    \caption{\gptfivetwo\ (with high reasoning effort) answers to the INDU questions. Each cell shows the number of correct answers across three repeats. The CT answers are referenced by R1 and R2. The CN and converse answers for R1 are shown as additional columns to the left of the table.}
    \label{fig:indu-gpt52}
\end{figure}

None of the LLMs tested gets all INDU questions correct (Fig. \ref{fig:accuracy-by-model-by-calculus}), but 15/33 models get all the INDU converse questions correct.
The best performing models are \gptfivetwo\ with high reasoning effort \repeats{660}{661}{664}{675} and \emph{Grok-4} (660/675).
Of the 40 incorrect \gptfivetwo\ answers
31 are CN and 9 CT. (Fig. \ref{fig:indu-gpt52}). All \gptfivetwo\ answers to the CT questions are answered reliably at least twice out of our three repeats, contributing to the \gptfivetwo\ overall benchmark score but further illustrating a lack of reliable reasoning.

Models sometimes hallucinate relation names (for example \emph{D>} is not a valid INDU relation, but
models use it as an answer on 174 occasions, including 42 by \emph{Gemini 2.5 Flash Lite} and even 27 by
\emph{GPT 5.1} (although never in the later OpenAI model \gptfivetwo). We also see similar problems
for \emph{S=} and \emph{D=} for example. These mistakes all indicate a lack of understanding of the relation terminology/semantics, e.g. that if one interval is during another it cannot be equal to or larger than the other one.

\subsection{STAR (9 relations, 99 questions)}

\begin{figure}
    \centering
    \includegraphics[width=0.5\textwidth]{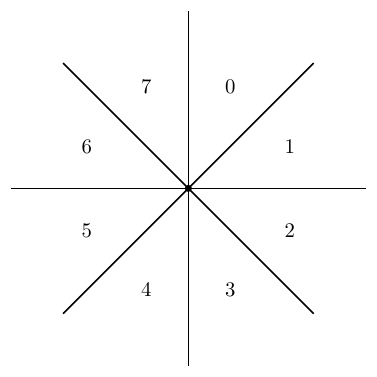}
    \caption{The eight regions, numbered 0 to 7, in the STAR calculus. ID is the identity relation (not shown explicitly but is the central point).}
    \label{fig:star}
\end{figure}

The Revised STAR (henceforth STAR) calculus is used for representing and reasoning about qualitative directions between pairs of points, x and y, in a two-dimensional Euclidean space with respect to a reference direction \cite{renz2004qualitative}; see Fig. \ref{fig:star}.


\gptfivetwo\ with high reasoning is the only model to
get all STAR questions correct (Fig. \ref{fig:accuracy-by-model-by-calculus}), but many models (22/33) get all the STAR converse questions correct and ten get all the STAR CN questions correct.


On one occasion, \kimi\ answers the question \emph{If 4(x,y) and 2(y,z) then what are the possible relations between x and z?} by trying to apply trigonometry; after a long reasoning chain mentioning sin, cos, and arctan, it just stops
without a valid answer. A total of (7/99) \kimi\ answers failed to give an answer with a valid STAR relation.

\subsection{9IM (8 relations, 80 questions)}

\begin{figure}
    \centering
    \includegraphics[width=\textwidth]{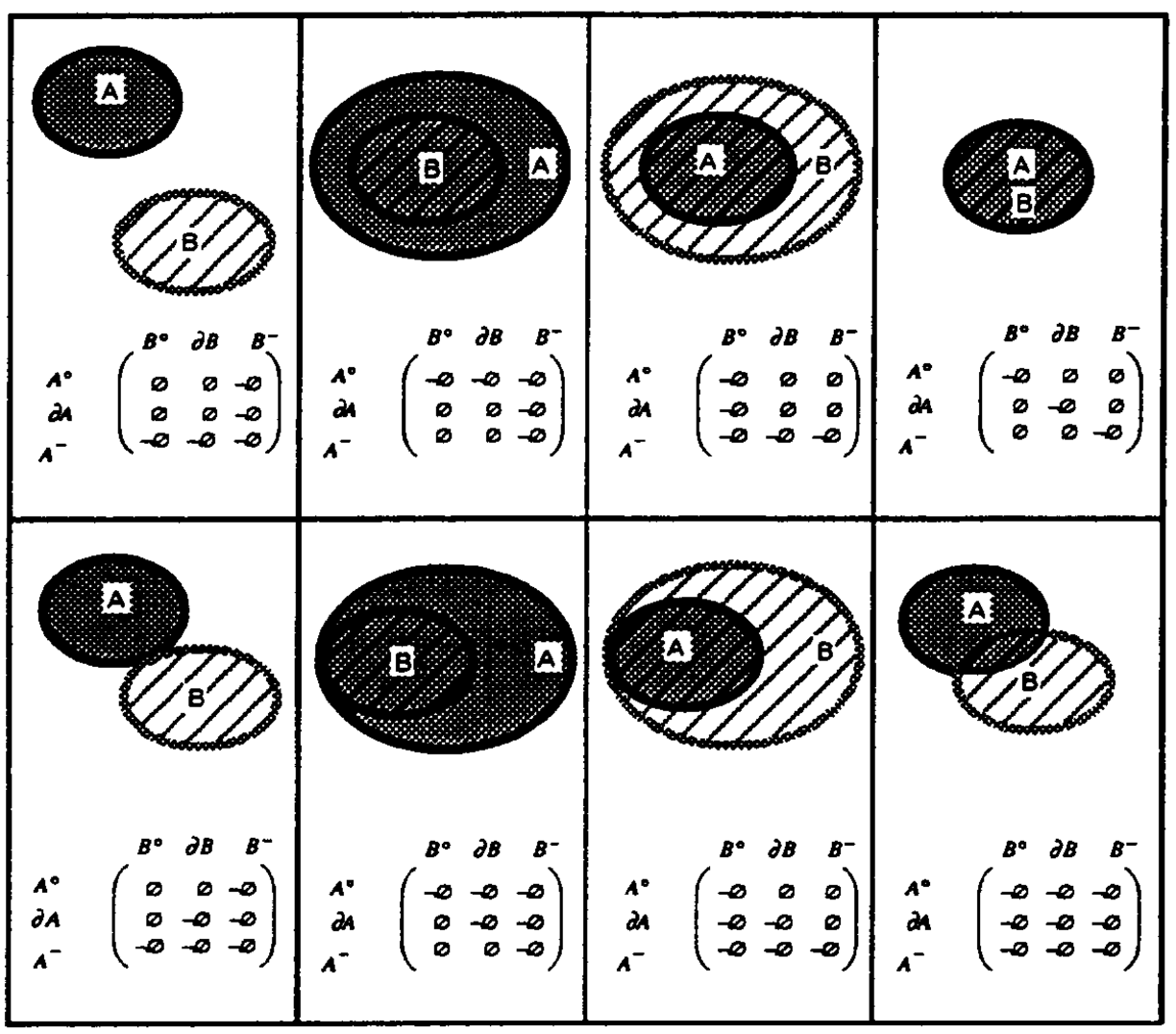}
    \caption{The nine intersection model (9IM, figure courtesy of Egenhofer \citep{egenhofer1994deriving}). The relations are
    Disjoint (D - similar to \DC), Contains (CT - similar to \NTPPi),
    Inside (I - similar to \NTPP), Equal (E - similar to EQ),
    Meets (M similar to \EC), Covers (CV similar to \TPPi), Covered By (CB - similar to \TPP),
    and Overlap (O - similar to \PO).}
    \label{fig:9im}
\end{figure}

The nine intersection model (9IM) is a qualitative spatial calculus for representing and reasoning about spatial relationships between one-piece non-empty regular closed regions with connected boundaries in a two-dimensional topological space \citep{egenhofer1994deriving,egenhofer1993critical,egenhofer19949}.
The relations are similar to RCC-8, but are described in terms of the nine pairwise intersections of interiors, boundaries and exteriors of two regions and considers whether these are empty or non-empty, resulting in eight jointly exhaustive and pairwise disjoint binary spatial relations (Fig. \ref{fig:9im}), once particular constraints on the regions involved are taken into account.

None of the LLMs tested gets all 9IM questions correct (Fig. \ref{fig:accuracy-by-model-by-calculus}), and only 9/33 models get all the 9IM converse questions correct.
Converse questions for 9IM were the least accurate of all the calculi -- we suggest that the the description of 9IM in terms of matrices can obscure understanding of converseness.

The best performing model for 9IM is \gptfivetwo\ with high reasoning, \repeats{77}{75}{76}{80}. Of the 12 incorrect answers, ten are CN\footnote{As mentioned above, that the notion of a CN differs in the 9IM compared to other calculi, and RCC-8 in particular.} and only two are CT.
The problematic CT questions are M(x,y) $\circ$ CV(y,z) (2/3) and CV(x,y) $\circ$ O(y,z) (2/3) but \gptfivetwo\ gets the corresponding RCC-8 compositions EC(x,y) $\circ$ TPPi(x,y) (3/3) and TPPi(x,y) $\circ$ PO(x,y) (3/3) correct.
The problematic CN questions are for M(x,y) (0/3), E(x,y) (2/3), CV(x,y) (0/3) and CB(x,y) (0/3) -- \gptfivetwo\ also has trouble with the similar RCC-8 relation CNs for  \EC\  (2/3) , \EQ\  (1/3), and \TPPi\  (1/3) but gets all \TPP\  CN answers correct (3/3).

\gptfivetwo\ always gets the CN for \emph{M} incorrect -- correctly specifying \emph{D}, but
missing \emph{O}. Similarly, \gptfivetwo\ always gets the CN for \emph{CB} incorrect, correctly specifying \emph{I} but missing \emph{O} and \emph{E}.

\emph{GPT-4o} answered \emph{EC} once (and referred to it as ``Externally Connects''), perhaps suggesting a confusion between RCC-8 and 9IM and a reliance on, and confusion in, the training data. Gemini 2.0 Flash once answered \emph{IB} rather \emph{I} in answer to \emph{If CB(x,y) then what is the relation between y and x?} -- \emph{IB} is not a valid 9IM relation name.

\subsection{RCC-22 (22 relations, 528 questions)}

\begin{figure}
    \centering
    \includegraphics[width=0.8\textwidth]{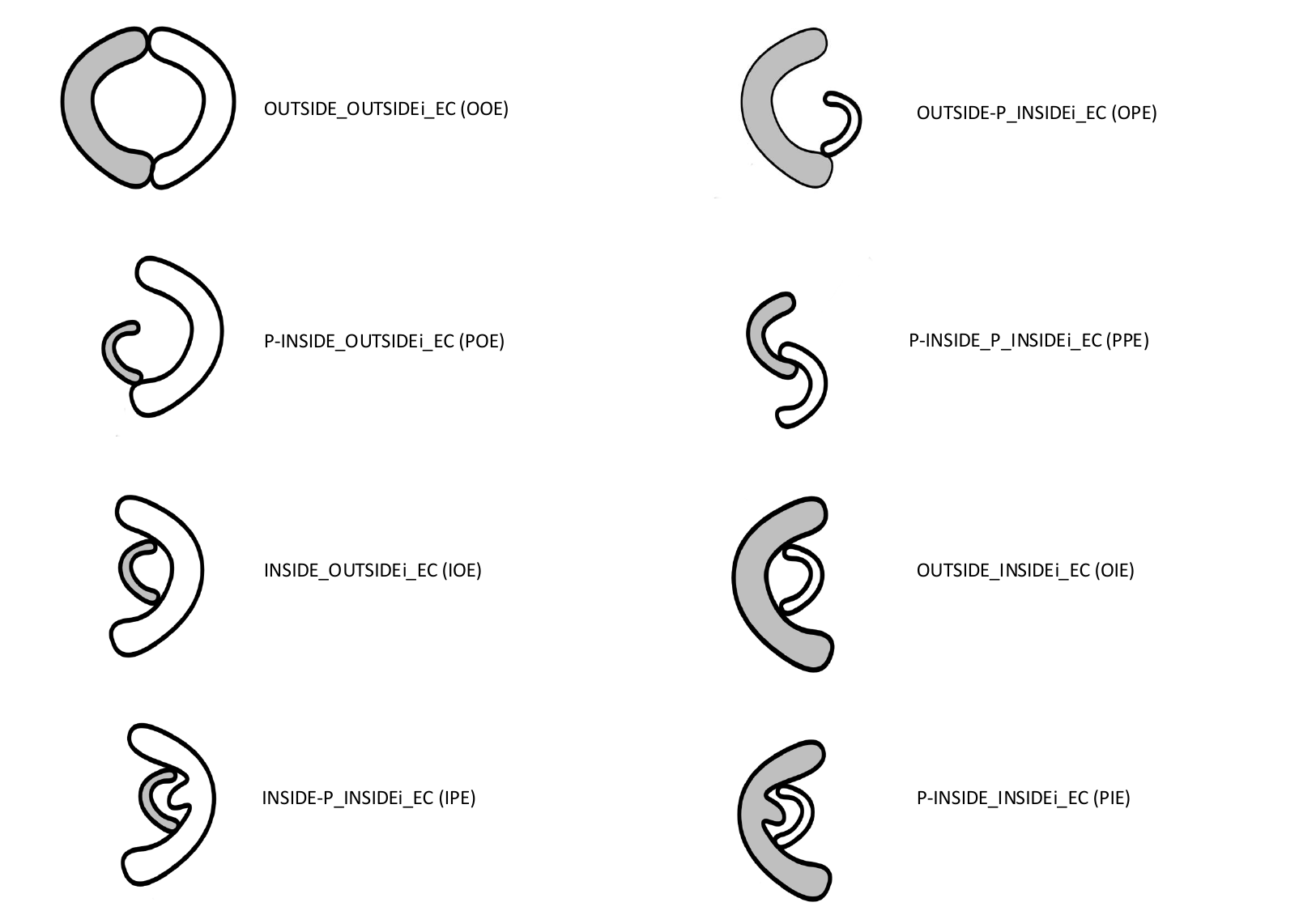}
    \caption{The RCC-22 \EC\   relations from \cite{cohn1997qualitative}. RCC-22 also includes \EQ, \PO, \TPP, \TPPi, \NTPP\ and \NTPPi\ as described for RCC-8. OUTSIDE\_OUTSIDEi\_DC (\OOD), P-INSIDE\_OUTSIDEi\_DC (\POD),
    INSIDE\_OUTSIDEi\_DC (\IOD), INSIDE-P\_INSIDEi\_DC (\IPD), P-INSIDE\_P-INSIDEi\_DC (\PPD), OUTSIDE-P\_INSIDEi\_DC (\OPD), OUTSIDE\_INSIDEi\_DC (\OID), and P-INSIDE\_INSIDEi\_DC (\PID) are similar to the above except that the
    regions are disconnected. Note that we use the long names in our LLM experiments but sometimes refer to the
    short names in brackets for brevity in this paper.
    A full list of the RCC-22 conceptual neighbours is given in Table \ref{tbl:rcc22cn}.
    }
    \label{fig:rcc22}
\end{figure}

The Region Connection Calculus, RCC-22 is a qualitative spatial calculus for representing and reasoning about spatial relationships between convex or concave, non-empty regular closed regions of uniform dimension in a topological space \citep{cui1993}; also see RCC-23 in \cite{cohn1997qualitative} and \cite{bennett1994spatial}. The \EC\   RCC-22 relations are illustrated in Fig. \ref{fig:rcc22}.  The \DC\  relations can be easily visualised from these depictions by simply moving the two regions slightly apart from each other.

\begin{figure}
    \centering
    \includegraphics[width=0.99\textwidth]{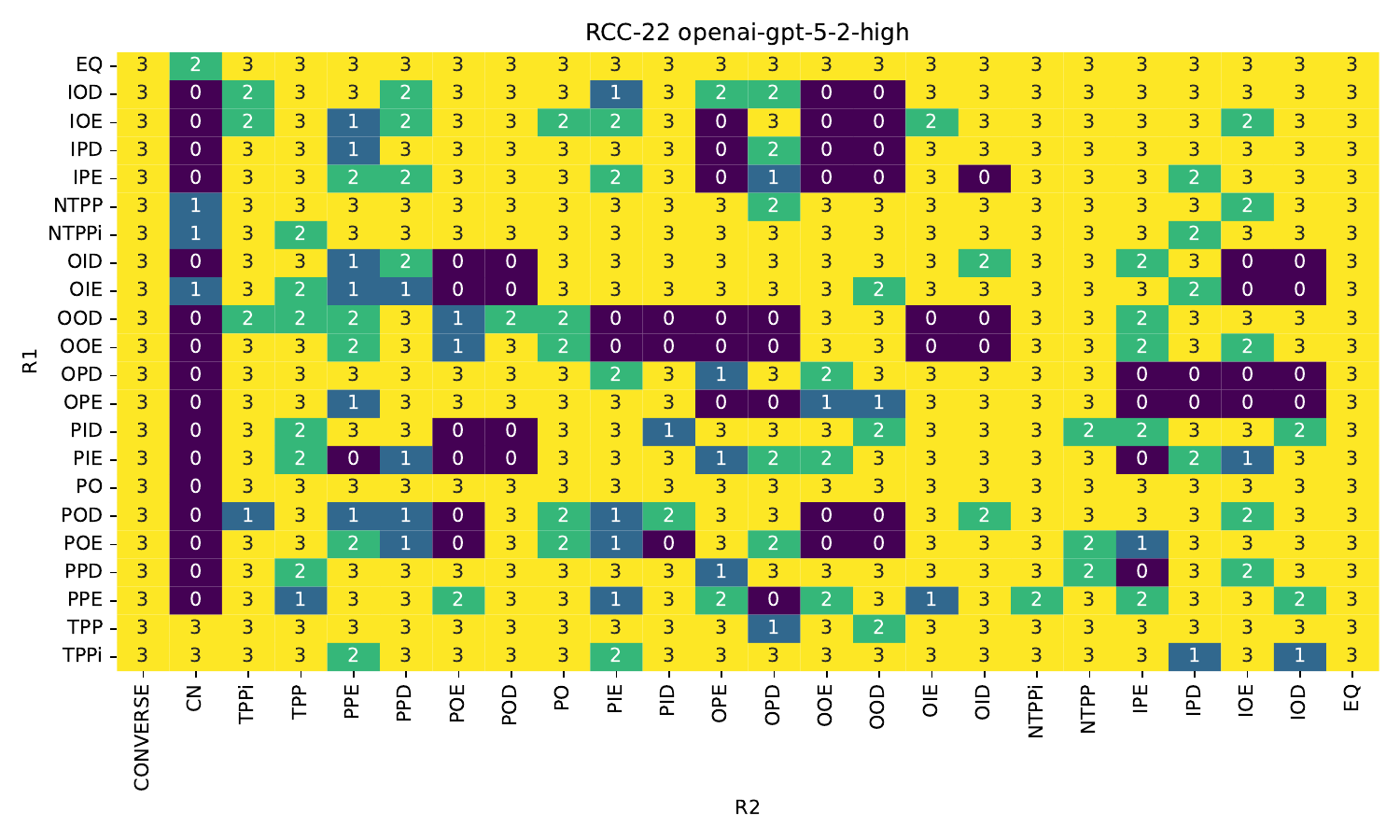}
    \caption{\gptfivetwo\ (with high reasoning effort) answers to the RCC-22 questions. Each cell shows the number of correct answers across three repeats. The CT answers are referenced by R1 and R2. The CN and converse answers for R1 are shown as additional columns to the left of the table.}
    \label{fig:rcc22-gpt52}
\end{figure}

None of the LLMs tested gets all RCC-22 questions correct (Fig. \ref{fig:accuracy-by-model-by-calculus}), but 9/33 models get all the RCC-22 converse questions correct.
Grok-4 is the best performing model (415/528),
but \gptfivetwo\ performs comparably  \repeats{407}{415}{408}{528}, (Fig. \ref{fig:rcc22-gpt52}).

\gptfivetwo\ always gets the CN for \TPPi\ and \TPP\ correct. It gets \EQ\ correct (2/3) and each of \NTPP,  \NTPPi, and \OIE\ correct 1/3 but never gets the CN for the other relations correct, suggesting that it struggles with the additional complexity of RCC-22 relations over the better known RCC-8 relations. Note that all of these instance have an `Outside' (O) component in at least one position, whereas compositions involving \PPD\, and to a lesser extent, \PPE\ are have fewer incorrect compositions.

Inspecting Fig. \ref{fig:rcc22-gpt52} it is clear that there are some relations which prove particularly problematic for \gptfivetwo: the composition of \OOD\ or \OOE\ with any of \PIE, \PID, \OPE, \OPD, \OIE, or \OID\ is incorrect across all three repetitions. Similarly, when R2 is  \OOD\ or \OOE, and  R1 is any of the dual relations of \PIE, \PID, \OPE, \OPD, \OIE, or \OID, i.e. any of \IOD, \IOE, \IPD, \IPE, \POD, or \POE, then all three repetitions are incorrect.


For both \emph{Grok-4} and \gptfivetwo\ with high reasoning effort, the most prevalent false negatives were all
RCC-22 specific relation names (not RCC-8 names), suggesting that
models have trouble grasping the additional complexity of relations
involving concave regions.

When testing RCC-22 with Gemini-Pro-2 we occasionally got an error response with a \texttt{finishReason} of \texttt{MAX\_TOKENS} and a \texttt{thoughtsTokenCount} of 65535. Although there was no response text, we surmise that the model had become stuck in
overthinking the question. Resubmitting the question\footnote{In this model it is not possible to set the temperature, and presumably the default is not 0, so a second attempt has a chance to avoid the original problem.} cleared the problem and the benchmark could resume. Similarly we could not get XAI Grok 4 to answer the question \emph{If INSIDE\_PINSIDEi\_DC(x,y) and PINSIDE\_INSIDEi\_DC(y,z) then what are the possible relations between x and z?} via OpenRouter, despite multiple retries, and so we used the XAI API directly. It is troubling that models seem to perform differently across APIs -- possibly affecting evaluation studies such as the one described here.

Clearly RCC-22 is the most complex of the calculi tested, with even \gptfivetwo\ with high reasoning
answering with RCC-8 relation names rather than RCC-22 relation names on 11 occasions.
\emph{GPT OSS 120b} once answered with \emph{PINSIDE\_INSII\_DC} (An invalid RCC-22 relation name),  again
suggesting possible problems with tokenisation. Further RCC-22 token analysis is included in \ref{supplementary}.

\subsection{Comparison of RCC-8 with 9IM.}

Although described using a different formalism, 9IM is similar to RCC-8, with relations roughly corresponding as follows:
\emph{D} (disjoint) -- \DC,
\emph{CT} (contains) -- \NTPPi,
\emph{I} (inside) -- \NTPP,
\emph{E} (equal) -- \EQ,
\emph{M} (meet) -- \EC,
\emph{CV} (covers) -- \TPPi,
\emph{CB} (coveredBy) -- \TPP, and
\emph{O} (overlap) -- \PO. The converse relations and CT are equivalent, but there is a slight difference in the CN: The CN is called closest topological distance in 9IM and does not include equivalents of  the links between \EQ\ -- \PO, \EQ\ -- \NTPP, or \EQ\ -- \NTPPi.

However, even \gptfivetwo\ with high reasoning effort makes different mistakes across the two (Fig. \ref{fig:rcc8v9im}); for CT reasoning there are only two incorrect answers across the three repeats for 9IM but  only one for RCC8. For the CN there are 10 incorrect answers for the 9IM (but three relations which are incorrect across all three repeats)  and 11 for RCC-8 (but only one relation which is completely incorrect across all three repeats). Arguably, computing the 9IM closest topological distance is the easier computation compared to the RCC-8 CN since it can be computed purely syntactically by comparing the entries in the two matrices, whereas the CN for RCC-8 requires actual \emph{spatial} reasoning -- from that viewpoint the fact that there are as many incorrect answers as there are for the 9IM is surprising.


\begin{figure}
    \centering
    \includegraphics[width=0.99\textwidth]{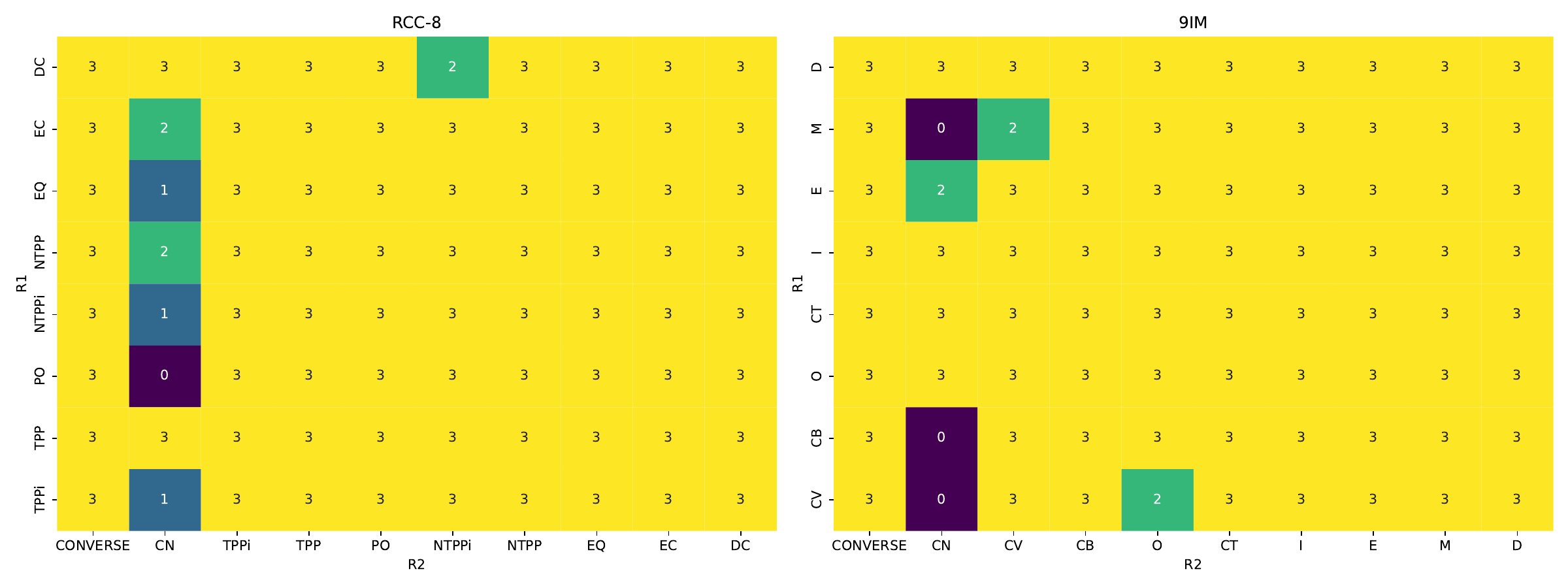}
    \caption{Comparison of \gptfivetwo\ with high reasoning effort answers to the RCC-8 questions (left) and 9IM question (right). Each cell shows the number of correct answers across three repeats. The CT answers are referenced by R1 and R2. The CN and converse answers for R1 are shown as additional columns to the left of each table.}
    \label{fig:rcc8v9im}
\end{figure}

\subsection{Extended experiments.}

Whether relation names are symbols, words or nonce words generally makes little difference to accuracy (Fig. \ref{fig:extras-o1}), though for PA  and RCC-8 \emph{Text Nonce Infix} is a bit better than \emph{Text Nonce Prefix}.
Interestingly, for RCC-22, then Nonce descriptions give the best accuracy, perhaps suggesting that when forced to reason rather than rely on any training data, the models actually do a bit better, but it is hard to be definitive as to the reason.
However, accuracy declines if calculi are described in terms of schematic diagrams rather than in natural language. In the case
of a schematic nonce prefix, the LLM is not told the name of the calculus or the names of the relations, (deliberately) hindering its
recourse to training data.


\begin{figure}
    \centering
    \includegraphics[width=0.8\textwidth]{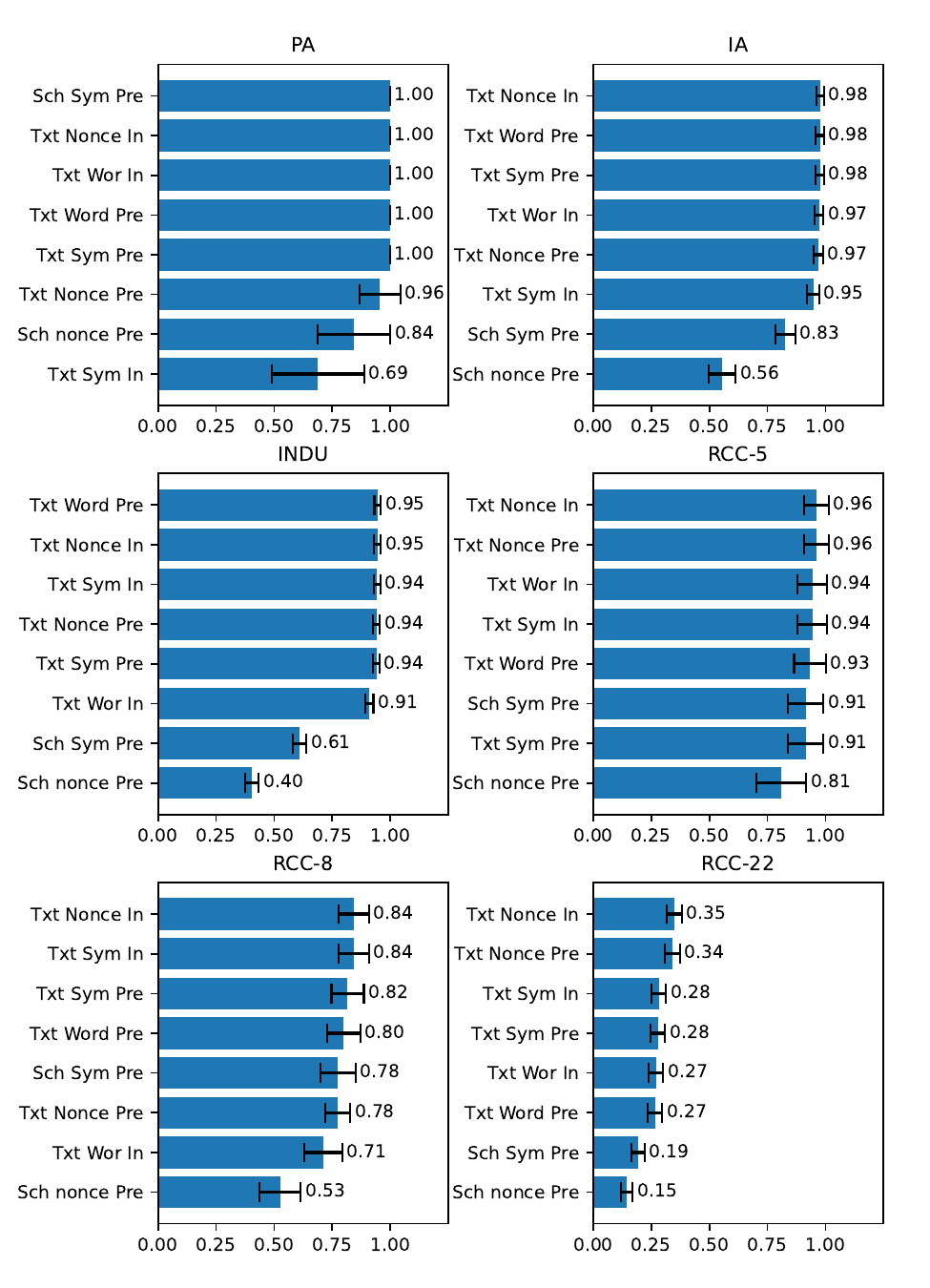}
    \caption{Accuracy by description style for o1  answers to PA, IA, INDU, RCC-5, RCC-8 and RCC-22 questions.
    The error bar is the prediction interval.}
    \label{fig:extras-o1}
\end{figure}

Whether relations use prefix or infix notation makes little difference to answers (Fig. \ref{fig:jaccard-o1}). However
answers are dissimilar when words, symbols and nonce words are compared, i.e. different mistakes are made across these differing prompting styles.

\begin{figure}
    \centering
    \includegraphics[width=\textwidth]{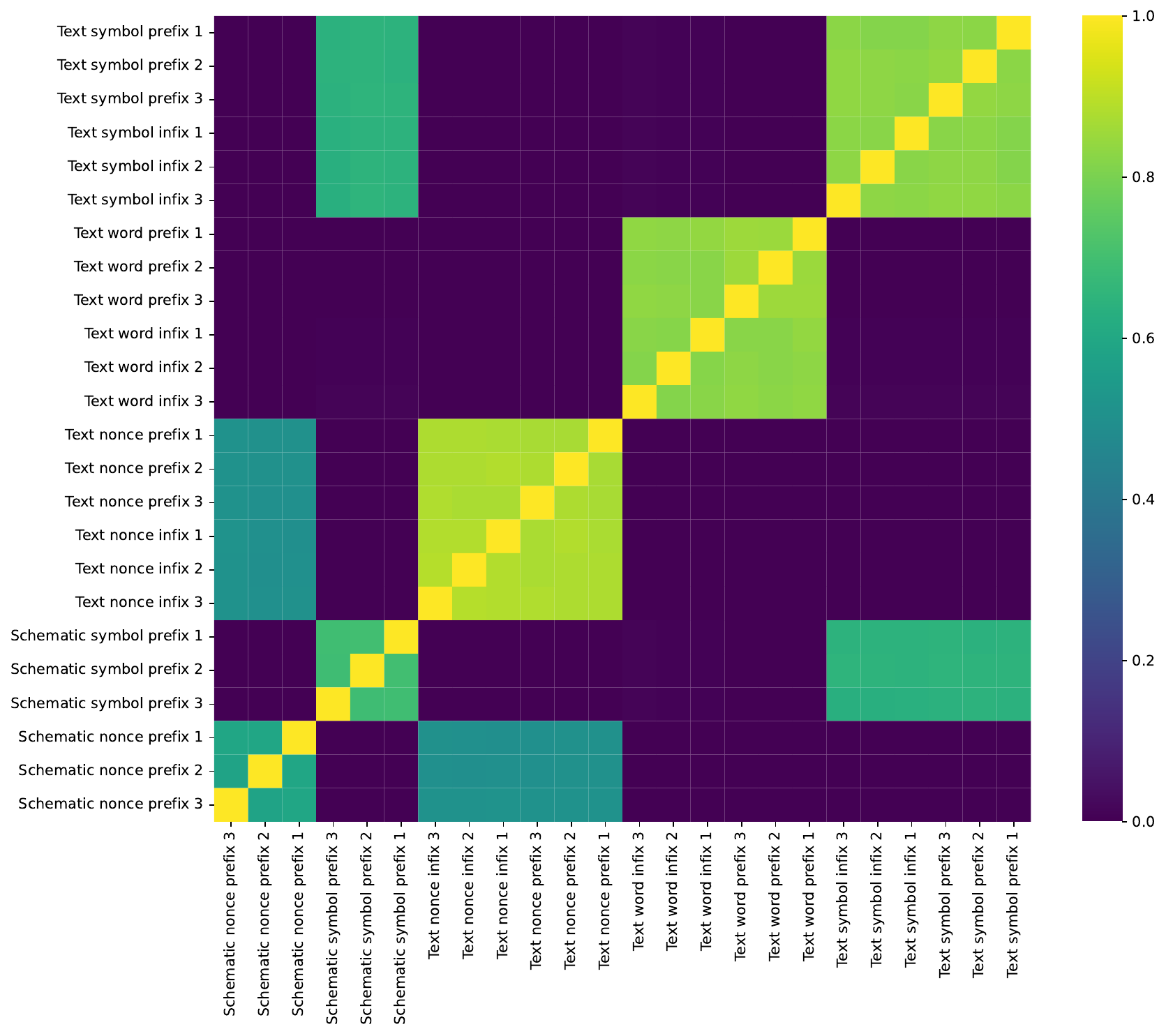}
    \caption{Mean Jaccard agreement between answers given by different description styles for o1  answers to PA, IA, INDU, RCC-5, RCC-8 and RCC-22 combined.}
    \label{fig:jaccard-o1}
\end{figure}

\subsection{RCC Narrowing to coarser calculi}

If we take the RCC-22 answers, we can ``collapse'' these to RCC-8 -- e.g. by converting all the \DC\   relations to plain \DC\   and all the \EC\   relations to plain \EC. We can then reassess the accuracy using a weighted average. Similarly we can collapse RCC-22 to RCC-5, and RCC-8 to RCC-5. For \gptfivetwo, whilst RCC-22 accuracy improves by narrowing, this is not the case for RCC-8 (For more information see \ref{supplementary}, Figures \ref{fig:rcc-coarse-accuracy}, \ref{fig:rcc-coarse-comparison}).

\subsection{Model costs}

o1 is the most expensive model used ($\$560$ per repeat) illustrating
the high costs of evaluating some frontier models.
The number of tokens used, and costs per repeat of the benchmark are included in \ref{supplementary}, Table \ref{tab:llm-costs}.
All models are presented with the same prompts, and so the varying number of input tokens shows
that tokenisation algorithms differ by models. \emph{Grok 4} uses the most fine-grained tokenizer and gpt-oss-20b uses the most coarse-grained tokenizer. \gptfivetwo\  with high reasoning effort costs \$77.28 per run for an accuracy of 92\% compared to \$49.82 with default reasoning effort with an accuracy of 89\% -- a 55\% increase in cost for 3\% performance gain.
\gptthreefive\   was the fastest model tested (median latency 0.3s) and \emph{deepseek-reasoner r1} the slowest (226s) suggesting
that models that undertake more reasoning incur a performance cost.
\emph{Gemini 2.5 flash-lite} consumed 83.6\% more  output tokens compared to the next most output token hungry model (\emph{deepseek-reasoner-r1}), and only 4\% more input tokens but has  much worse accuracy (0.36 vs 0.72), so more effort at  inference time does not necessarily compensate for a smaller model\footnote{Neither model gives an  official size, but the  former is given as 671B, and the latter is estimated online as under 50B.}.

\subsection{Token counts}

\gptfivetwo\ with high reasoning effort uses an order of magnitude more tokens to solve
RCC-22 questions than RCC-8 questions, and two orders of magnitude more tokens than RCC-5
(Fig. \ref{fig:token_counts}). Although the order of calculi is not the same as the order for accuracy (Fig. \ref{fig:accuracy-by-calculus}), it is further evidence that RCC-22 is the most difficult calculus for LLMs, and PA the easiest.

\begin{figure}[!t]
    \centering
    \includegraphics[width=0.8\textwidth]{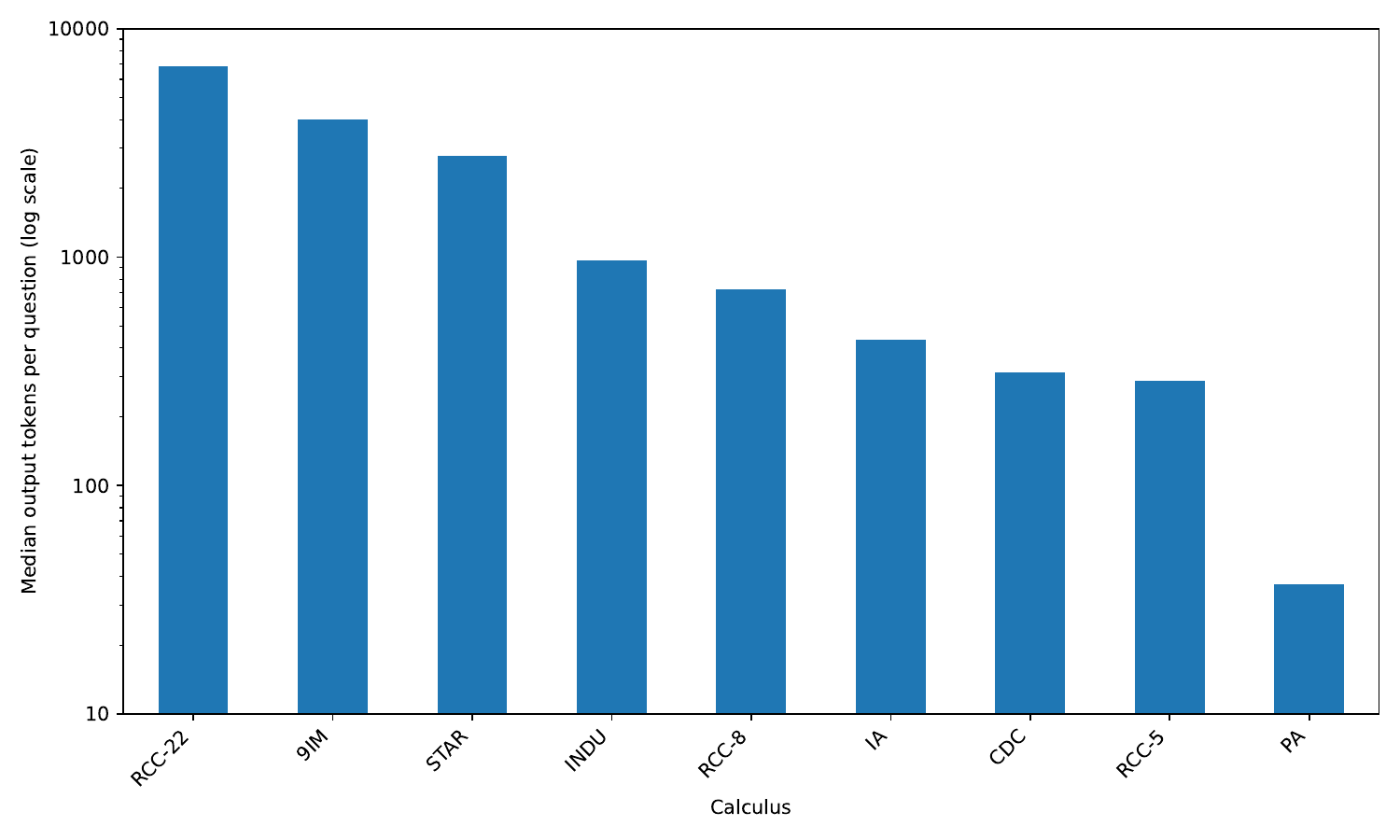}
    \caption{Median output token counts per question by calculus for \gptfivetwo\ with high reasoning effort (log scale).}
    \label{fig:token_counts}
\end{figure}

\section{Summary}
\subsection{Contributions}

\noindent We summarise the main contributions of this paper:
\begin{itemize}
\item Benchmark related contributions
\begin{itemize}
    \item We publish an open-source \qstrsmall\ benchmark data set comprising 1806 questions and answers, covering nine QSTR calculi. We use this to test a range of contemporary LLMs on key qualitative spatial relation  inference tasks.

\item We also publish an open-source \qstrlarge\ benchmark comprising 14372 questions and answers designed to probe LLM reasoning capabilities more deeply.

\item In both these data sets we vary description and question styles for each canonical question in order to test model robustness, reliability and true reasoning capabilities.

\item Both benchmarks are correct by construction thus eliminating possible errors in the ground truth.

\item Our benchmark provides a framework which can be easily extended to a much larger number of QSTR calculi if desired, or as and when new calculi become available.
\item The RCC-22 composition table is included here for the first time.
\end{itemize}

\item Experiment related
\begin{itemize}

\item We use these two datasets to conduct an extensive evaluation of a representative set of nine QSTR calculi, across three inference tasks, using multiple prompt variations and present a detailed analysis of the results.
\item We go beyond model comparison, using fine-grained analysis (e.g. token counts and reasoning traces) to gain insight into the reasoning process.

\item We show that even the best models make mistakes, and moreover are not consistent across repeats; the smallest models perform very poorly in general, providing a challenge for deployment on consumer/edge hardware.

\item We contribute to the question of whether LLMs actually reason, or simply retrieve answers from their training data via pattern matching, in particular through the use of schematic, nonce and swapped prompting styles.

\item We publish all our results including all HTTP request-response pairs for all models\footnote{As recommended by Burnell et al. \citep{burnell2023rethink}}, providing full traceability, and a resource for future study.
\end{itemize}
\end{itemize}
\subsection{Conclusions}

Whilst LLMs demonstrate a capacity for reasoning  -- or at least simulating reasoning (all models perform better than guessing alone), none of the frontier models tested is able to correctly answer all QSTR benchmark questions. LLM reasoning is non-deterministic and inconsistent -- the same question presented to the same LLM on different repeats, can result in different answers (even with temperature set to 0 and a fixed random seed, where available).

The extended RCC-8 experiment, where we vary the prompt considerably provides evidence that models are able to generalise and call upon both training data and information given in the prompt.

RCC-22 has fewer base relations (22) than INDU (25) but is more challenging for the LLMs tested.  RCC-22 is not widely documented (except in respect to the closely related RCC-23 calculus, which in turn is also not that widely documented or discussed) and the CN of both RCC-22 and RCC-23  have never been published, however INDU has 124 citations, some of which include explicit descriptions. The low accuracy seen in schematic, nonce experiments (where the calculus is not named and the relations are not described conventionally) provides some evidence of models' reliance on training data, and at the same time apparent difficulties in interpreting the schematic diagrams which are readily understandable by humans.

Models have become more accurate over time, but  also in general more expensive, and have higher latency. The high latency of large reasoning models could make their use impractical in real time systems. A logic program written in a language such as  Prolog could give fast, consistent, accurate and reliable results to QSTR questions, but would be unable to cope with different natural language phrasing. We note that
neurosymbolic approaches aim to use LLMs for autoformalisation -- translating problems expressed in natural language, into a syntax suitable for an external solver (e.g. \cite{li2024advancing,mcpheat2025decompsrdatasetdecomposedanalyses}, though such approaches are still subject to the vagaries of inaccurate/hallucinated translations by the LLM into the symbolic form which becomes increasingly problematic the larger the statement of the problem since just a single mis-translation will cause the symbolic reasoner to fail.

Although the domains of STAR and CDC are similar, and RCC-8 and 9IM are similar, results are different, suggesting that  models are not able to generalise across representations (Fig. \ref{fig:accuracy-by-calculus}).

Models sometimes give answers that include hallucinated (confabulated?) relation names -- it is one thing to be incorrect,
but quite another to invent a new syntax to describe an answer\footnote{Leyton-Brown and Shoham \citep{leyton2024understanding} suggest that one measure of understanding in an AI system is whether it gives \emph{ridiculous} answers rather than merely incorrect ones; a hallucinated relation name arguably counts as the former.}.

Although there is much interest in small, and so-called edge language models \citep{belcak2025smalllanguagemodelsfuture, zheng2025review}, our results suggest that, at the time of writing, larger language models are much more accurate at reasoning about QSTR calculi.

Finally, we note that models seem to fundamentally misunderstand the idea of identity (EQ) relationships, which ought to be one of the easiest relationships to reason about (and indeed as noted above, Ragni et al \citep{ragni2007cross} omitted EQ from their tests with human subjects as they assumed it would be trivial). Similarly, many models cannot reason reliably about parthood relationships, which again would seem to be one of the more straightforward relationships.

\section{Limitations}
\label{sec:limitations}
There are inevitably a number of limitations to the presented work.  These fall into two main categories: benchmark-related and experiment-related. We list the most important limitations below.

{\bf Benchmark-related}
Rather than cover every single QSTR known to us in the literature, we selected what we regard as a representative subset.  The reasons for this were given in section \ref{sec:exp-design}; nevertheless, particularly on the spatial side, adding more calculi would allow for a more comprehensive benchmark; moreover, part of our selection criteria was to choose better known (more highly cited) calculi as being of more interest -- but these might be better covered by the LLMs' training data and thus potentially less of a good test of any actual reasoning ability -- selecting lesser known calculi would not only increase coverage overall, but would also mean that the models might be less well trained on such calculi.

It is important to note that this benchmark is not a set of completely independent questions.  Within the set of prompts for any model, there are many shared relation names, and except for the different description varieties (e.g. infix, prefix...), there are only three kinds of questions (converse, CT, CN). Thus if a model is performing well on some CT questions for example, this might mean that there is a prior probability that it will perform well on other CT questions. Having a wider set of possible spatial reasoning tasks would reduce this effect, but our aim here was specifically to test the abilities of models on the three ``standard'' inference tasks in the literature (converse, CT, CN).

Although, particularly in the extended dataset, we have a variety of different ways of describing the calculi (prefix, infix, schematic...) for each of these there is only a single linguistic phrasing -- providing more linguistic variety would provide a greater test of LLM abilities.

{\bf Experiment-related}
As noted above (and see Fig. \ref{tab:llm-costs}), we already incurred considerable costs, even if much of this was offset by our sponsorship from Microsoft and Google Deepmind.  Further experiments would have allowed at least three repeats for all models and questions. Especially where we were only able to afford a single repeat this limits the statistical significance of the comparisons of that model against others, so these results should be taken as indicative rather than any stronger claim about relative model performance.

Related to this is the question of what metric should be used to measure performance; in section \ref{sec:exp-design} we argued for micro-averages of accuracies.  In earlier work \cite{cohn2024largelanguagemodelsreason} we had used the Jaccard Index which is more lenient since it gives credit for slightly wrong answers (e.g. an extra relation or a missing relation in the CT and CN experiments).  However we now believe that is too generous -- if LLMs are to be a replacement for symbolic reasoning (the GQR and SparQ would give 100\% accuracy on all the questions in the benchmark, provided an oracle could translate into the required symbolic description required by the reasoner), then we should hold an LLM to the same standard.  However, if desired, Jaccard Index results could easily be computed from the stored experimental results in the GitHub repository without any further LLM computation. Other possible metrics which we have not given here include a task-balance macro-average where the averages for each of the tasks (converse, CT, CN) are computed individually; this would have the effect of downweighting the CT results (since there are quadratically many of them compared to the linear number of converse and CN questions).  Giving a calculus-balanced macro-average might also be interesting, i.e. computing the average for each calculus individually, and then averaging the results; this would give PA the same weighting in the overall average compared to INDU which has many more questions in it.

These results do not take into account the number of tokens used for each relation symbol. For example, RCC-8 uses a mean of 1.75 tokens per relation (\DC, \EC, \PO, and \EQ\ use only one token, but \NTPP\ and \NTPPi\ use three tokens\footnote{According to OpenAI's tiktoken GPT-5.x tokenizer, \url{https://github.com/openai/tiktoken}, retrieved May 2026.}). However, the RCC-22 \PPE\ relation uses nine tokens. It may be that tokenisation complexity hinders reasoning and reduces performance for some calculi and relation combinations.

\section{Future work}
\label{sec:future-work}
Although the evaluation presented here is extensive, it is certainly not exhaustive (and indeed could never be given the ever changing landscape of LLMs). Many opportunities for further research exist, and we briefly discuss some of these here.

Firstly, we fully expect AI systems to improve and for new LLMs to be made available. Our \qstrsmall\ benchmark provides an opportunity to evaluate new AI systems and compare their performance to the models tested herein. Indeed, the QSTR performance of smaller models might be improved with fine tuning.

More QSTR calculi could be tested -- some of these are already listed in section \ref{sec:exp-design}. It would also be interesting to translate QSTR questions to other languages (especially low resource languages) and compare LLM performance (we have already noted above some work in comparing spatial reasoning across English, Hindi and Swedish\citep{mcpheat2025decompsrdatasetdecomposedanalyses}).

Some calculi can be used with algebraic operations (e.g. \cite{vilain1986constraint} show addition and multiplication with the temporal point-based calculi), but we have not yet tested LLM ability to perform these operations.

Moreover, since spatial reasoning in humans is fundamentally grounded in the real 2D or 3D world, we would like to test multi-modal, diagrammatic representations of calculi with Vision Language Models, and indeed we have already started this line of work\citep{cosit-26-diagrammatic}. Related to this is the idea of embedding spatial testing in real-world scenarios rather than abstract ``x'', ``y'' examples; this already exists to some extent in some existing benchmarks (e.g. \citep{li2024reframing}).

Other forms of reasoning could also be explored, for example consistency checking -- there is extensive work on finding maximal tractable subsets of QSTR calculi (e.g. \citep{drakengren1997complete,renz1999complexity}) for use when determining whether a knowledge base of facts expressed using disjunctions of relations is consistent or not (in general determining consistency in such calculi and situations is at least NP-complete).  Can LLMs determine consistency of such knowledge bases and how do they fare in particular on instances which are at the phase transition between easy and hard problems?

\section{Acknowledgements}

This work was supported by the Fundamental Research priority area of The Alan Turing Institute, by the Economic and
Social Research Council (ESRC) under grant ES/W003473/1, by the EPSRC under grant EP/Z003512/1, and by the Special
Funds of Tongji University for the ``Sino-German Cooperation 2.0 Strategy''. We thank Microsoft Research -- Accelerating Foundation Models Research program, for the provision of Azure resources to access OpenAI models, and Google Deepmind for the provision of free LLM credits for some of the Gemini model experiments, without which many of the experiments in this paper would not have been possible. We also thank David Randell for helpful discussions about the RCC-22 conceptual neighbourhood.

\section{Data Access Statement}
All data used in this work are either publicly available already (in particular the CTs and CNs to be found in SPARQ \citep{wolter2009sparq} and GQR \citep{gantner2008gqr}) or were generated specifically for this research and can be found in an accompanying GitHub repository\footnote{\url{https://github.com/RobBlackwell/QSTRBench} accessed May 2026.}.

\section{Author Contributions}
AGC originally conceived the idea for the paper and wrote an earlier arXiv paper focussing just on RCC-8 and ChatGPT. REB conducted all the experimental work and wrote the initial draft of the paper. Both authors co-designed the experiments and analysed the results.

\section{Ethics}
There are no ethical considerations in the work described here, except perhaps for the carbon cost of conducting the experiments which is endemic to the important task of evaluating foundation models.

\section{Declaration of generative AI and AI-assisted technologies in the manuscript preparation process}

During the preparation of this work the authors used OpenAI ChatGPT and Microsoft Copilot in order to check and refine the writing. After using these tools/services, the authors reviewed and edited the content as needed and take full responsibility for the content of the published article.


\bibliographystyle{elsarticle-num-names}
\bibliography{references}






\pagebreak
\appendix
\section{Example prompts for RCC-8}
\label{example-prompts}

\subsection{Text symbol prefix}

You are a helpful assistant who answers questions about qualitative spatial and temporal calculi.

The Region Connection Calculus (RCC-8) is a qualitative spatial calculus for representing and reasoning about spatial relationships between non-empty regular closed regions of uniform dimension in a topological space. It consists of eight jointly exhaustive and pairwise disjoint binary spatial relations.

DC(x,y) means that x and y are disconnected and do not have intersecting boundaries.

EQ(x,y) means that x and y are coincident.

PO(x,y) means that x and y have a region z in common but neither is part of the other.

TPP(x,y) means that the boundaries of x and y intersect, x and y are not coincident, and x is a part of y.

EC(x,y) means that x and y have intersecting boundaries but do not share any interior parts.

NTPP(x,y) means that the boundaries of x and y do not intersect, x and y are not coincident, and x is a part of y.

TPPi(x,y) means that the boundaries of x and y intersect, x and y are not coincident, and y is a part of x.

NTPPi(x,y) means that the boundaries of x and y do not intersect, x and y are not coincident, and y is a part of x.

I will now ask you a question about these relations. There may be more than one possible relation, in which case name all of the possible answers. Answer the question and provide the final answer in the form: "\#\#\# Answer:". The final answer should only contain relations separated by semicolon with no extraneous text.

If DC(x,y) and DC(y,z) then what are the possible relations between x and z?

\subsection{Text word prefix}

You are a helpful assistant who answers questions about qualitative spatial and temporal calculi.

The Region Connection Calculus (RCC-8) is a qualitative spatial calculus for representing and reasoning about spatial relationships between non-empty regular closed regions of uniform dimension in a topological space. It consists of eight jointly exhaustive and pairwise disjoint binary spatial relations.

disconnected(x,y) means that x and y are disconnected and do not have intersecting boundaries.

equal(x,y) means that x and y are coincident.

partial\_overlap(x,y) means that x and y have a region z in common but neither is part of the other.

tangential\_proper\_part(x,y) means that the boundaries of x and y intersect, x and y are not coincident, and x is a part of y.

externally\_connected(x,y) means that x and y have intersecting boundaries but do not share any interior parts.

nontangential\_proper\_part(x,y) means that the boundaries of x and y do not intersect, x and y are not coincident, and x is a part of y.

tangential\_proper\_part\_inverse(x,y) means that the boundaries of x and y intersect, x and y are not coincident, and y is a part of x.

nontangential\_proper\_part\_inverse(x,y) means that the boundaries of x and y do not intersect, x and y are not coincident, and y is a part of x.

I will now ask you a question about these relations. There may be more than one possible relation, in which case name all of the possible answers. Answer the question and provide the final answer in the form: "\#\#\# Answer:". The final answer should only contain relations separated by semicolon with no extraneous text.

If disconnected(x,y) and disconnected(y,z) then what are the possible relations between x and z?

\subsection{Text symbol infix}

You are a helpful assistant who answers questions about qualitative spatial and temporal calculi.

The Region Connection Calculus (RCC-8) is a qualitative spatial calculus for representing and reasoning about spatial relationships between non-empty regular closed regions of uniform dimension in a topological space. It consists of eight jointly exhaustive and pairwise disjoint binary spatial relations.

x DC y means that x and y are disconnected and do not have intersecting boundaries.

x EQ y means that x and y are coincident.

x PO y means that x and y have a region z in common but neither is part of the other.

x TPP y means that the boundaries of x and y intersect, x and y are not coincident, and x is a part of y.

x EC y means that x and y have intersecting boundaries but do not share any interior parts.

x NTPP y means that the boundaries of x and y do not intersect, x and y are not coincident, and x is a part of y.

x TPPi y means that the boundaries of x and y intersect, x and y are not coincident, and y is a part of x.

x NTPPi y means that the boundaries of x and y do not intersect, x and y are not coincident, and y is a part of x.

I will now ask you a question about these relations. There may be more than one possible relation, in which case name all of the possible answers. Answer the question and provide the final answer in the form: "\#\#\# Answer:". The final answer should only contain relations separated by semicolon with no extraneous text.

If x DC y and y DC z then what are the possible relations between x and z?

\subsection{Text word infix}

You are a helpful assistant who answers questions about qualitative spatial and temporal calculi.

The Region Connection Calculus (RCC-8) is a qualitative spatial calculus for representing and reasoning about spatial relationships between non-empty regular closed regions of uniform dimension in a topological space. It consists of eight jointly exhaustive and pairwise disjoint binary spatial relations.

x disconnected y means that x and y are disconnected and do not have intersecting boundaries.

x equal y means that x and y are coincident.

x partial\_overlap y means that x and y have a region z in common but neither is part of the other.

x tangential\_proper\_part y means that the boundaries of x and y intersect, x and y are not coincident, and x is a part of y.

x externally\_connected y means that x and y have intersecting boundaries but do not share any interior parts.

x nontangential\_proper\_part y means that the boundaries of x and y do not intersect, x and y are not coincident, and x is a part of y.

x tangential\_proper\_part\_inverse y means that the boundaries of x and y intersect, x and y are not coincident, and y is a part of x.

x nontangential\_proper\_part\_inverse y means that the boundaries of x and y do not intersect, x and y are not coincident, and y is a part of x.

I will now ask you a question about these relations. There may be more than one possible relation, in which case name all of the possible answers. Answer the question and provide the final answer in the form: "\#\#\# Answer:". The final answer should only contain relations separated by semicolon with no extraneous text.

If x disconnected y and y disconnected z then what are the possible relations between x and z?

\subsection{Text nonce prefix}

You are a helpful assistant who answers questions about qualitative spatial and temporal calculi.

The calculus of interest in this question is a qualitative spatial calculus for representing and reasoning about spatial relationships between non-empty regular closed regions of uniform dimension in a topological space. It consists of eight jointly exhaustive and pairwise disjoint binary spatial relations.

asharge(x,y) means that x and y are disconnected and do not have intersecting boundaries.

extrave(x,y) means that x and y are coincident.

therine(x,y) means that x and y have a region z in common but neither is part of the other.

mustide(x,y) means that the boundaries of x and y intersect, x and y are not coincident, and x is a part of y.

veniend(x,y) means that x and y have intersecting boundaries but do not share any interior parts.

delizzy(x,y) means that the boundaries of x and y do not intersect, x and y are not coincident, and x is a part of y.

influed(x,y) means that the boundaries of x and y intersect, x and y are not coincident, and y is a part of x.

sistion(x,y) means that the boundaries of x and y do not intersect, x and y are not coincident, and y is a part of x.

I will now ask you a question about these relations. There may be more than one possible relation, in which case name all of the possible answers. Answer the question and provide the final answer in the form: "\#\#\# Answer:". The final answer should only contain relations separated by semicolon with no extraneous text.

If asharge(x,y) and asharge(y,z) then what are the possible relations between x and z?

\subsection{Text nonce infix}

You are a helpful assistant who answers questions about qualitative spatial and temporal calculi.

The calculus of interest in this question is a qualitative spatial calculus for representing and reasoning about spatial relationships between non-empty regular closed regions of uniform dimension in a topological space. It consists of eight jointly exhaustive and pairwise disjoint binary spatial relations.

x asharge y means that x and y are disconnected and do not have intersecting boundaries.

x extrave y means that x and y are coincident.

x therine y means that x and y have a region z in common but neither is part of the other.

x mustide y means that the boundaries of x and y intersect, x and y are not coincident, and x is a part of y.

x veniend y means that x and y have intersecting boundaries but do not share any interior parts.

x delizzy y means that the boundaries of x and y do not intersect, x and y are not coincident, and x is a part of y.

x influed y means that the boundaries of x and y intersect, x and y are not coincident, and y is a part of x.

x sistion y means that the boundaries of x and y do not intersect, x and y are not coincident, and y is a part of x.

I will now ask you a question about these relations. There may be more than one possible relation, in which case name all of the possible answers. Answer the question and provide the final answer in the form: "\#\#\# Answer:". The final answer should only contain relations separated by semicolon with no extraneous text.

If x asharge y and y asharge z then what are the possible relations between x and z?

\subsection{Schematic symbol prefix}

You are a helpful assistant who answers questions about qualitative spatial and temporal calculi.

The Region Connection Calculus (RCC-8) is a qualitative spatial calculus for representing and reasoning about spatial relationships between non-empty regular closed regions of uniform dimension in a topological space. It consists of eight jointly exhaustive and pairwise disjoint binary spatial relations. These relations are illustrated in one dimensional space below. A capital X means that that space is occupied by the region x only, a capital Y means that that space is occupied by the region y only, a capital B means that both x and y occupy that space, and a tilde indicates that neither x nor y occupy that space.

DC(x,y) is illustrated by the following text-based schematics

XXXX{\textasciitilde}YYYY or YYYY{\textasciitilde}XXXX

EQ(x,y) is illustrated by the following text-based schematic

BBBB

PO(x,y) is illustrated by the following text-based schematic

XXBBYY

TPP(x,y) is illustrated by the following text-based schematics

BBYY or YYBB

EC(x,y) is illustrated by the following text-based schematic

XXXXYYYY

NTPP(x,y) is illustrated by the following text-based schematic

YBBY

TPPi(x,y) is illustrated by the following text-based schematics

BBXX or XXBB

NTPPi(x,y) is illustrated by the following text-based schematic

XBBX

I will now ask you a question about these relations. There may be more than one possible relation, in which case name all of the possible answers. Answer the question and provide the final answer in the form: "\#\#\# Answer:". The final answer should only contain relations separated by semicolon with no extraneous text.

If DC(x,y) and DC(y,z) then what are the possible relations between x and z?

\subsection{Schematic nonce prefix}

You are a helpful assistant who answers questions about qualitative spatial and temporal calculi.

The calculus of interest in this question is a qualitative spatial calculus for representing and reasoning about spatial relationships between non-empty regular closed regions of uniform dimension in a topological space. It consists of eight jointly exhaustive and pairwise disjoint binary spatial relations. These relations are illustrated in one dimensional space below. A capital X means that that space is occupied by the region x only, a capital Y means that that space is occupied by the region y only, a capital B means that both x and y occupy that space, and a tilde indicates that neither x nor y occupy that space.

asharge(x,y) is illustrated by the following text-based schematics

XXXX{\textasciitilde}YYYY or YYYY{\textasciitilde}XXXX

extrave(x,y) is illustrated by the following text-based schematic

BBBB

therine(x,y) is illustrated by the following text-based schematic

XXBBYY

mustide(x,y) is illustrated by the following text-based schematics

BBYY or YYBB

veniend(x,y) is illustrated by the following text-based schematic

XXXXYYYY

delizzy(x,y) is illustrated by the following text-based schematic

YBBY

influed(x,y) is illustrated by the following text-based schematics

BBXX or XXBB

sistion(x,y) is illustrated by the following text-based schematic

XBBX

I will now ask you a question about these relations. There may be more than one possible relation, in which case name all of the possible answers. Answer the question and provide the final answer in the form: "\#\#\# Answer:". The final answer should only contain relations separated by semicolon with no extraneous text.

If asharge(x,y) and asharge(y,z) then what are the possible relations between x and z?

\subsection{Point symbol prefix }

You are a helpful assistant who answers questions about qualitative spatial and temporal calculi.

The Region Connection Calculus (RCC-8) is a qualitative spatial calculus for representing and reasoning about spatial relationships between non-empty regular closed regions of uniform dimension in a topological space. It consists of eight jointly exhaustive and pairwise disjoint binary spatial relations.

DC(x,y) means that the closures of x and y have no points in common.

EQ(x,y) means that all the points in the closure of x are also in the closure of y and all the points in the closure of y are also in the closure of x.

PO(x,y) means that x and y share at least one interior point, the interior of x contains at least one point not in the interior of y, and the interior of y contains at least one point not in the interior of x.

TPP(x,y) means that all points in the closure of x are also in the closure of y, the interior of y contains at least one point not in the interior of x, there is a region z whose closure shares no interior points with x or y, and there is at least one point common to the closures of x, y and z.

EC(x,y) means that the closures of x and y have points in common, but their interiors do not.

NTPP(x,y) means that all points in the closure of x are also in the closure of y, there is at least one point in the interior of y which is not in the interior of x, and for any region z which shares no interior points with the interior of y, then the closure of z does not share any boundary points with the closure of x.

TPPi(x,y) means that the interior of x contains at least one point not in the interior of y, all points in the closure of y are also in the closure of x, there is a region z whose closure shares no interior points with x or y, and there is at least one point common to the closures of x, y and z.

NTPPi(x,y) means that there is at least one point in the interior of x which is not in the interior of y, all points in the closure of y are also in the closure of x, and for any region z which shares no interior points with the interior of x, then the closure of z does not share any boundary points with the closure of y.

I will now ask you a question about these relations. There may be more than one possible relation, in which case name all of the possible answers. Answer the question and provide the final answer in the form: "\#\#\# Answer:". The final answer should only contain relations separated by semicolon with no extraneous text.

If DC(x,y) and DC(y,z) then what are the possible relations between x and z?

\subsection{LaTeX symbol prefix}

You are a helpful assistant who answers questions about qualitative spatial and temporal calculi.

The Region Connection Calculus (RCC-8) is a qualitative spatial calculus for representing and reasoning about spatial relationships between non-empty regular closed regions of uniform dimension in a topological space. It consists of eight jointly exhaustive and pairwise disjoint binary spatial relations defined in terms of a primitive  topological connection relation C(x,y), whose intended interpretation is that the closures of regions x and y share at least one point.

The axiomatisation of C(x,y) is given as follows:

\textbackslash ( \textbackslash forall x C(x,x) \textbackslash )

\textbackslash ( \textbackslash forall x \textbackslash forall y \textbackslash left[ C(x,y) \textbackslash rightarrow C(y,x) \textbackslash right]\textbackslash )

Three auxiliary predicates are defined as follows:

\textbackslash ( P(x,y) \textbackslash equiv \textbackslash forall z \textbackslash left[ C(z,x) \textbackslash rightarrow C(z,y) \textbackslash right] \textbackslash )

\textbackslash ( O(x,y) \textbackslash equiv \textbackslash exists z \textbackslash left[ P(z,x) \textbackslash land P(z,y) \textbackslash right] \textbackslash )

\textbackslash ( PP(x,y) \textbackslash equiv \textbackslash left[ P(x,y) \textbackslash land \textbackslash lnot P(y,x) \textbackslash right] \textbackslash )

The eight relations are defined as follows:

\textbackslash ( DC(x,y) \textbackslash equiv \textbackslash lnot C(x,y) \textbackslash )

\textbackslash ( EQ(x,y) \textbackslash equiv \textbackslash left[ P(x,y) \textbackslash land P(y,x) \textbackslash right] \textbackslash )

\textbackslash ( PO(x,y) \textbackslash equiv \textbackslash left[ O(x,y) \textbackslash land \textbackslash neg P(x,y) \textbackslash land \textbackslash neg P(y,x) \textbackslash right] \textbackslash )

\textbackslash ( TPP(x,y) \textbackslash equiv \textbackslash left[ PP(x,y) \textbackslash land \textbackslash exists z \textbackslash left[ EC(z,x) \textbackslash land EC(z,y) \textbackslash right] \textbackslash right] \textbackslash )

\textbackslash ( EC(x,y) \textbackslash equiv \textbackslash left[ C(x,y) \textbackslash land \textbackslash neg O(x,y) \textbackslash right] \textbackslash )

\textbackslash ( NTPP(x,y) \textbackslash equiv \textbackslash left[ PP(x,y) \textbackslash land \textbackslash neg \textbackslash exists z \textbackslash left[ EC(z,x) \textbackslash land EC(z,y) \textbackslash right] \textbackslash right] \textbackslash )

\textbackslash ( TPP\_i(x,y)  \textbackslash equiv \textbackslash left[ PP(y,x) \textbackslash land \textbackslash exists z \textbackslash left[ EC(z,x) \textbackslash land EC(z,y) \textbackslash right] \textbackslash right] \textbackslash )

\textbackslash ( NTPP\_i(x,y) \textbackslash equiv \textbackslash left[ PP(y,x) \textbackslash land \textbackslash neg \textbackslash exists z \textbackslash left[ EC(z,x) \textbackslash land EC(z,y) \textbackslash right] \textbackslash right] \textbackslash )

I will now ask you a question about these relations. There may be more than one possible relation, in which case name all of the possible answers. Answer the question and provide the final answer in the form: "\#\#\# Answer:". The final answer should only contain relations separated by semicolon with no extraneous text.

If DC(x,y) and DC(y,z) then what are the possible relations between x and z?

\subsection{Unicode symbol prefix}

You are a helpful assistant who answers questions about qualitative spatial and temporal calculi.

The Region Connection Calculus (RCC-8) is a qualitative spatial calculus for representing and reasoning about spatial relationships between non-empty regular closed regions of uniform dimension in a topological space. It consists of eight jointly exhaustive and pairwise disjoint binary spatial relations defined in terms of a primitive  topological connection relation C(x,y), whose intended interpretation is that the closures of regions x and y share at least one point.

The axiomatisation of C(x,y) is given as follows:

$\forall$xC(x,x)

$\forall$x$\forall$y[C(x,y) $\to$ C(y,x)]

Three auxiliary predicates are defined as follows:

P(x,y) $\equiv$ $\forall$z[C(z,x) $\to$ C(z,y)]

O(x,y) $\equiv$ $\exists$z[P(z,x) $\wedge$ P(z,y)]

PP(x,y) $\equiv$ [P(x,y) $\wedge$ $\neg$P(y,x)]

The eight relations are defined as follows:

DC(x,y) $\equiv$ $\neg$C(x,y)

EQ(x,y) $\equiv$ [P(x,y) $\wedge$ P(y,x)]

PO(x,y) $\equiv$ [O(x,y) $\wedge$ $\neg$P(x,y) $\wedge$ $\neg$P(y,x)]

TPP(x,y) $\equiv$ [PP(x,y) $\wedge$ $\exists$z[EC(z,x) $\wedge$ EC(z,y)]]

EC(x,y) $\equiv$ [C(x,y) $\wedge$ $\neg$O(x,y)]

NTPP(x,y) $\equiv$ [PP(x,y) $\wedge$ $\neg$$\exists$z[EC(z,x) $\wedge$ EC(z,y)]]

TPPi(x,y) $\equiv$ [PP(y,x) $\wedge$ $\exists$z[EC(z,x) $\wedge$ EC(z,y)]]

NTPPi(x,y) $\equiv$ [PP(y,x) $\wedge$ $\neg$$\exists$z[EC(z,x) $\wedge$ EC(z,y)]]

I will now ask you a question about these relations. There may be more than one possible relation, in which case name all of the possible answers. Answer the question and provide the final answer in the form: "\#\#\# Answer:". The final answer should only contain relations separated by semicolon with no extraneous text.

If DC(x,y) and DC(y,z) then what are the possible relations between x and z?

\subsection{Reverse symbol prefix}

You are a helpful assistant who answers questions about qualitative spatial and temporal calculi.

The Region Connection Calculus (RCC-8) is a qualitative spatial calculus for representing and reasoning about spatial relationships between non-empty regular closed regions of uniform dimension in a topological space. It consists of eight jointly exhaustive and pairwise disjoint binary spatial relations.

NTPPi(x,y) means that the boundaries of x and y do not intersect, x and y are not coincident, and y is a part of x.

TPPi(x,y) means that the boundaries of x and y intersect, x and y are not coincident, and y is a part of x.

NTPP(x,y) means that the boundaries of x and y do not intersect, x and y are not coincident, and x is a part of y.

EC(x,y) means that x and y have intersecting boundaries but do not share any interior parts.

TPP(x,y) means that the boundaries of x and y intersect, x and y are not coincident, and x is a part of y.

PO(x,y) means that x and y have a region z in common but neither is part of the other.

EQ(x,y) means that x and y are coincident.

DC(x,y) means that x and y are disconnected and do not have intersecting boundaries.

I will now ask you a question about these relations. There may be more than one possible relation, in which case name all of the possible answers. Answer the question and provide the final answer in the form: "\#\#\# Answer:". The final answer should only contain relations separated by semicolon with no extraneous text.

If DC(x,y) and DC(y,z) then what are the possible relations between x and z?

\subsection{None symbol prefix}

You are a helpful assistant who answers questions about qualitative spatial and temporal calculi.

The Region Connection Calculus (RCC-8) is a qualitative spatial calculus for representing and reasoning about spatial relationships between non-empty regular closed regions of uniform dimension in a topological space. It consists of eight jointly exhaustive and pairwise disjoint binary spatial relations.

I will now ask you a question about these relations. There may be more than one possible relation, in which case name all of the possible answers. Answer the question and provide the final answer in the form: "\#\#\# Answer:". The final answer should only contain relations separated by semicolon with no extraneous text.

If DC(x,y) and DC(y,z) then what are the possible relations between x and z?

\subsection{Text (swapped symbols) prefix}

You are a helpful assistant who answers questions about qualitative spatial and temporal calculi.

The calculus of interest in this question is a qualitative spatial calculus for representing and reasoning about spatial relationships between non-empty regular closed regions of uniform dimension in a topological space. It consists of eight jointly exhaustive and pairwise disjoint binary spatial relations.

PO(x,y) means that x and y are disconnected and do not have intersecting boundaries.

EC(x,y) means that x and y are coincident.

DC(x,y) means that x and y have a region z in common but neither is part of the other.

NTPPi(x,y) means that the boundaries of x and y intersect, x and y are not coincident, and x is a part of y.

EQ(x,y) means that x and y have intersecting boundaries but do not share any interior parts.

TPPi(x,y) means that the boundaries of x and y do not intersect, x and y are not coincident, and x is a part of y.

NTPP(x,y) means that the boundaries of x and y intersect, x and y are not coincident, and y is a part of x.

TPP(x,y) means that the boundaries of x and y do not intersect, x and y are not coincident, and y is a part of x.

I will now ask you a question about these relations. There may be more than one possible relation, in which case name all of the possible answers. Answer the question and provide the final answer in the form: "\#\#\# Answer:". The final answer should only contain relations separated by semicolon with no extraneous text.

If PO(x,y) and PO(y,z) then what are the possible relations between x and z?

\subsection{CT Preferences}

You are a helpful assistant who answers questions about qualitative spatial and temporal calculi.

The Region Connection Calculus (RCC-8) is a qualitative spatial calculus for representing and reasoning about spatial relationships between non-empty regular closed regions of uniform dimension in a topological space. It consists of eight jointly exhaustive and pairwise disjoint binary spatial relations.

DC(x,y) means that x and y are disconnected and do not have intersecting boundaries.

EQ(x,y) means that x and y are coincident.

PO(x,y) means that x and y have a region z in common but neither is part of the other.

TPP(x,y) means that the boundaries of x and y intersect, x and y are not coincident, and x is a part of y.

EC(x,y) means that x and y have intersecting boundaries but do not share any interior parts.

NTPP(x,y) means that the boundaries of x and y do not intersect, x and y are not coincident, and x is a part of y.

TPPi(x,y) means that the boundaries of x and y intersect, x and y are not coincident, and y is a part of x.

NTPPi(x,y) means that the boundaries of x and y do not intersect, x and y are not coincident, and y is a part of x.

I will now ask you a question about these relations. There may be more than one possible relation that might hold;in these cases just give me your single most preferred relation. Answer the question and provide the final answer in the form: "\#\#\# Answer:"

If DC(x,y) and DC(y,z) then what is your preferred relation between x and z?

\pagebreak
\section{Supplementary Information}
\label{supplementary}

\begin{figure}[!htbp]
    \centering
    \includegraphics[width=0.8\textwidth]{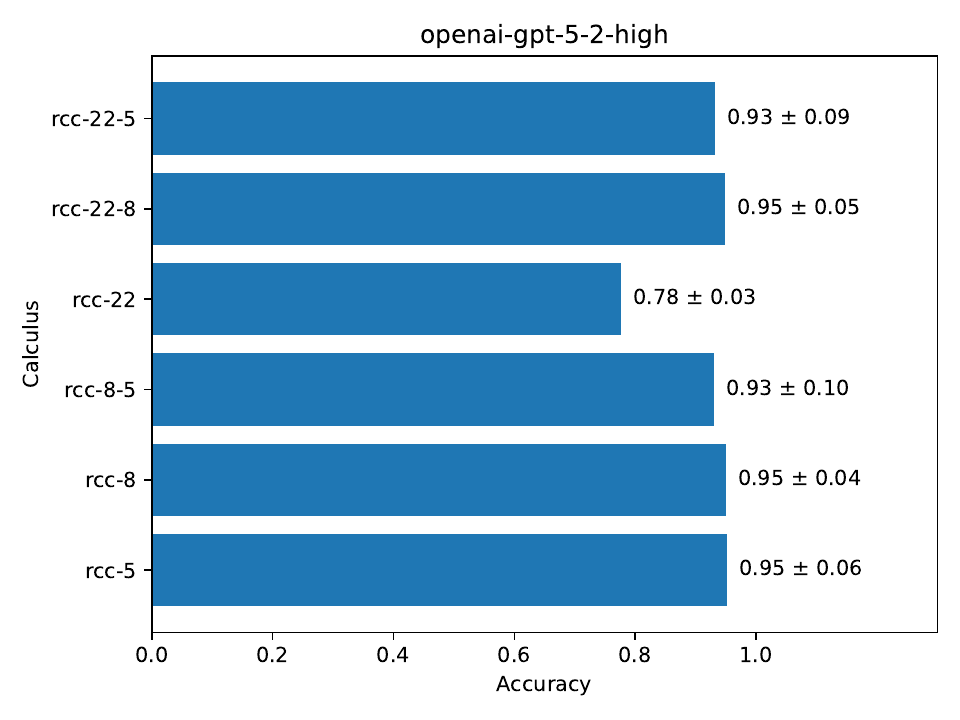}
    \caption{Accuracy of RCC-5, RCC-8 and RCC-22 compared to RCC-8 converted to RCC-5 and RCC-22 converted to RCC-8 and RCC-5 for \gptfivetwo\ with high reasoning effort.}
    \label{fig:rcc-coarse-accuracy}
\end{figure}

\begin{figure}
    \centering
    \includegraphics[width=0.8\textwidth]{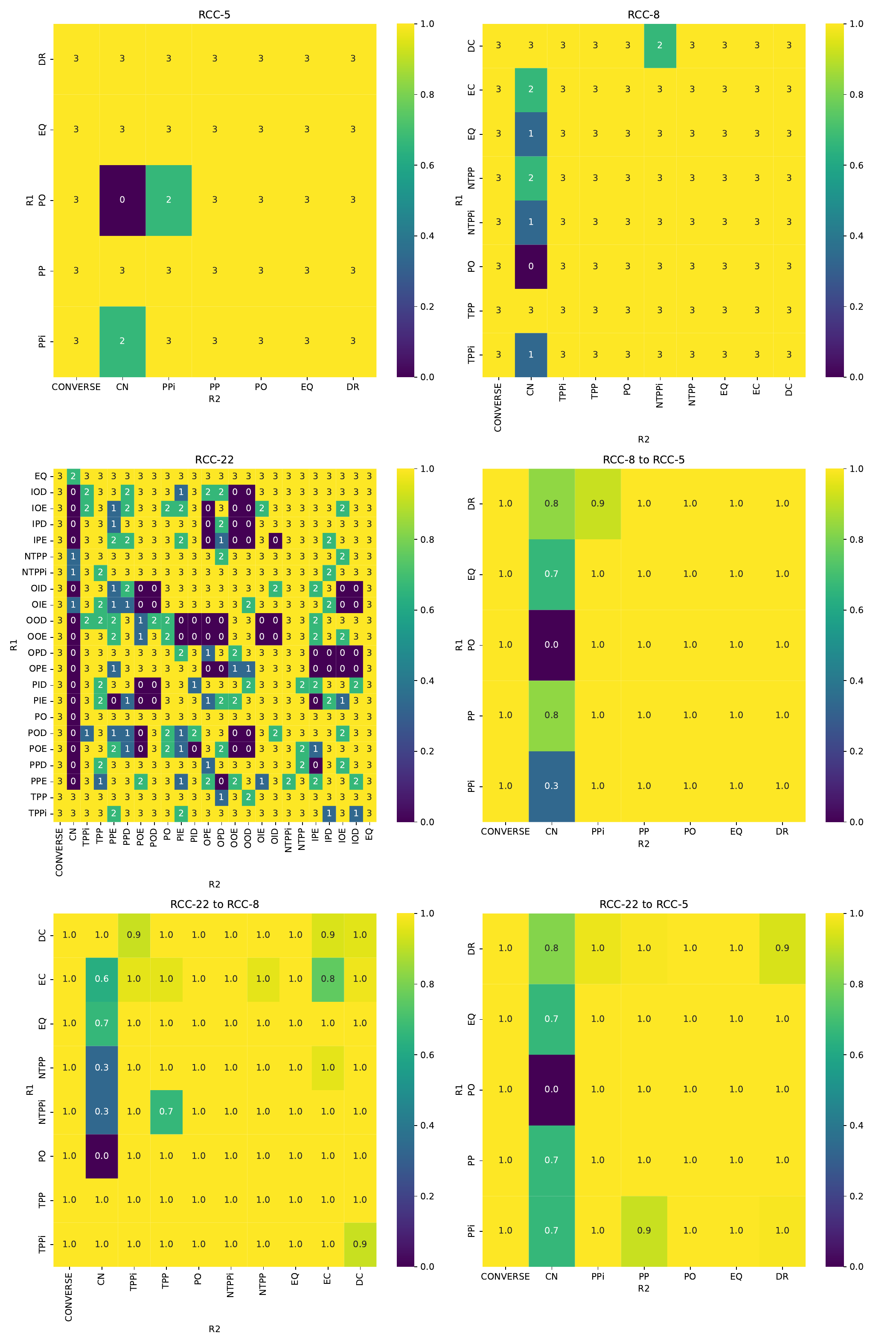}
    \caption{RCC-5, RCC-8 and RCC-22 compared to RCC-8 converted to RCC-5 and RCC-22 converted to RCC-8 and RCC-5 for for \gptfivetwo\ with high reasoning effort.}
    \label{fig:rcc-coarse-comparison}
\end{figure}

\begin{table}[!htbp]
\centering
\caption{LLM input and output tokens and costs per repeat for the \qstrsmall\ experiment (1806 questions). Pricing as at January 2026 (most experiments were run well before this date. All experimental data in the accompanying GitHub repository include time and date stamps). Note that the number of input tokens varies because models use different tokenization algorithms.}
\vspace{0.5em} 
\begin{tabular}{lrrr}
\toprule
model & input tokens & output tokens & total price \\
\midrule
azure-o1-2024-12-17 & 1,811,664 & 8,878,883 & \$559.91 \\
openai-gpt-5-2-high & 1,823,402 & 5,292,192 & \$77.28 \\
openai-gpt-5 & 1,811,664 & 7,492,504 & \$77.19 \\
openai-o3-2025-04-16 & 1,811,664 & 8,140,508 & \$68.75 \\
openai-gpt-5-2 & 1,811,769 & 3,331,990 & \$49.82 \\
deepseek-reasoner-r1 & 1,820,447 & 12,414,042 & \$32.31 \\
deepseek-r1-0528 & 1,652,641 & 11,410,840 & \$25.28 \\
azure-o4-mini-2025-04-16 & 1,811,664 & 5,105,107 & \$24.46 \\
claude-sonnet-4-20250514 & 2,020,095 & 916,392 & \$19.81 \\
claude-3-7-sonnet-20250219 & 2,020,095 & 794,236 & \$17.97 \\
gemini-2-5-pro & 1,894,769 & 1,383,616 & \$16.20 \\
glm-4-5 & 1,793,583 & 8,531,005 & \$13.85 \\
xai-grok-4 & 3,004,260 & 44,508 & \$9.68 \\
gemini-2-5-flash-lite & 1,894,769 & 22,792,650 & \$6.98 \\
kimi-k2 & 1,836,535 & 2,331,657 & \$6.51 \\
grok-4-1-fast & 2,050,614 & 11,848,285 & \$6.33 \\
gemini-2-5-flash & 1,875,678 & 1,785,390 & \$5.03 \\
deepseek-chat-v3-1 & 1,819,585 & 2,040,798 & \$4.45 \\
amazon-nova-pro-v1 & 1,845,629 & 591,883 & \$3.37 \\
openai-gpt-5-1 & 1,811,664 & 43,520 & \$2.70 \\
gpt-oss-20b & 1,780,792 & 9,451,797 & \$0.98 \\
gpt-oss-120b & 1,918,411 & 3,169,322 & \$0.68 \\
gemini-2.0-flash-001 & 1,876,336 & 926,260 & \$0.56 \\
xai-grok-3-mini & 1,781,598 & 34,427 & \$0.55 \\
llama-3-3-70b-instruct & 1,809,833 & 702,680 & \$0.24 \\
deepseek-v-3-2 & 1,818,810 & 4,342,283 & - \\
azure-gpt-4o-2024-11-20 & 1,813,470 & 578,540 & - \\
azure-gpt-45-preview-2025-02-07 & 1,813,470 & 1,278,731 & - \\
azure-gpt-35-turbo-0125 & 1,803,252 & 63,931 & - \\
\midrule
\textbf{Total} & 54,638,164 & 135,717,981 & \textbf{\$1,030.89} \\
\bottomrule
\bottomrule
\end{tabular}
\label{tab:llm-costs}
\end{table}

The RCC-22 relations that are common to RCC-8 (\EQ, \NTPP, \NTPPi, \TPP, \TPPi, and \PO) all tend to use fewer
tokens than the RCC-22 specific relations (Fig. \ref{fig:rcc-22-tokens}). This suggests, perhaps unsurprisingly,  that the RCC-22
specific relations are harder to reason about.

\begin{figure}
    \centering
    \includegraphics[width=0.98\textwidth]{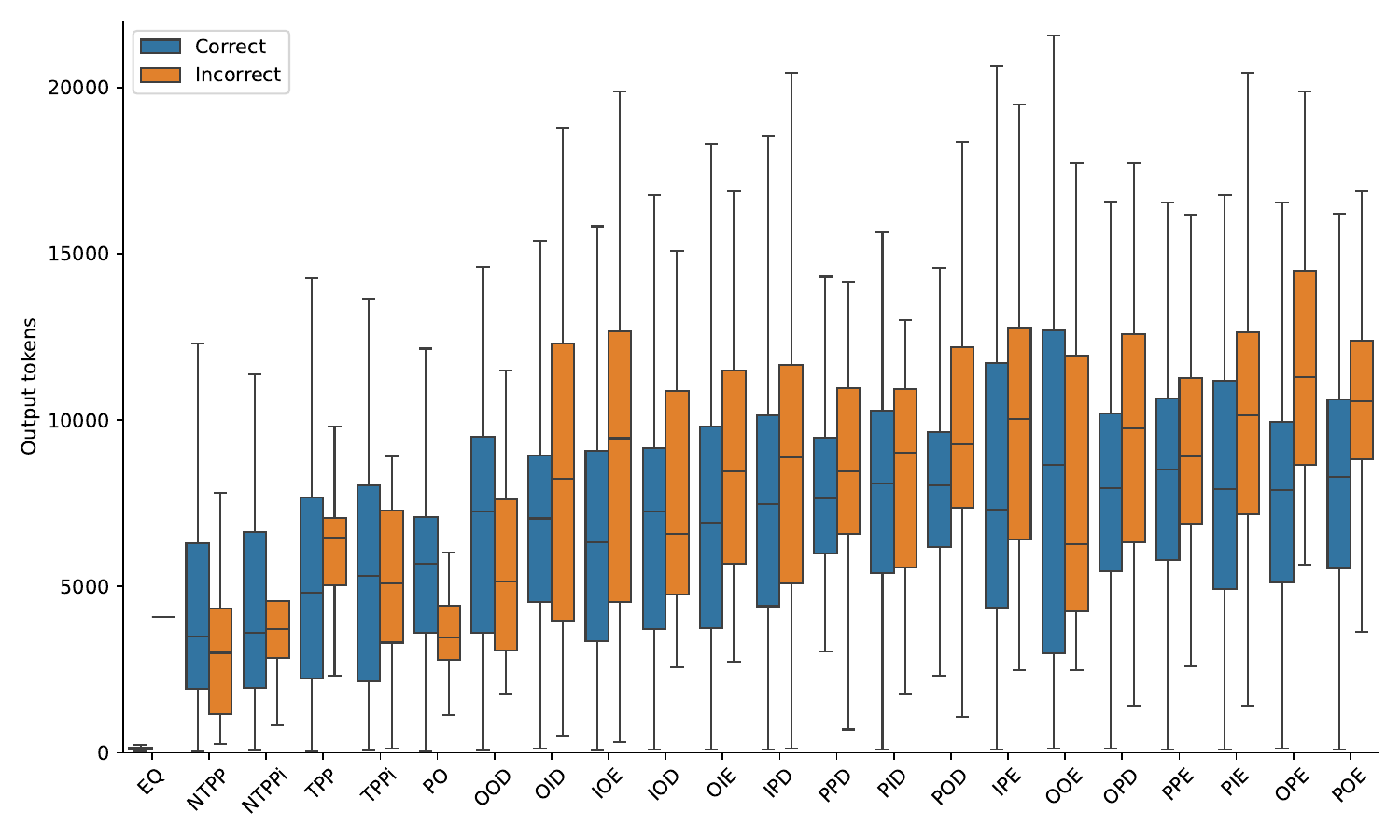}
    \caption{Number of tokens used by \gptfivetwo\ with high reasoning effort by RCC-22 relation appearing somewhere in the question (micro-average across all CT, CN and converse questions). Boxes indicate the interquartile range (IQR) with the central line denoting the median, while whiskers extend to 1.5 times IQR; outliers beyond this range are omitted. Hue indicates whether responses were correct. This further suggests that RCC-22 specific relations are harder to reason about than RCC-8, but beware that sample sizes differ in each of the bars.}
    \label{fig:rcc-22-tokens}
\end{figure}

\pagebreak

\section{Prolog program to compute the RCC-22 conceptual neighbourhood}
\label{sec:prolog-program}
RCC-22 is a finer grained version of RCC-8 with EC and DC subdivided into relations that also consider convexity and concavity.
For the subset of RCC-22 where the relation of interest is one of the original RCC-8 relations (i.e. \EQ, \TPP, \TPPi, \NTPP, \NTPPi\ and \PO)  then the conceptual neighbours of that relation are exactly the same as in the RCC-8 case, but noting that \PO\  also has neighbours which are all the refinements of \EC\   relation (i.e. relations of the form $\alpha$\_$\beta$\_\EC.
The computation of the relations between the RCC-22 specific relations is more complex.  Obviously any relation of the form $\alpha$\_$\beta$\_\EC\   has a neighbour of the form $\alpha$\_$\beta$\_\DC.
The question is which other relations are neighbours?  Underlying this computation is the notion that there are four regions of interest:  $x$ and its convex hull, $conv(x)$ and similarly for $y$ and its convex hull, $conv(y)$. The key relationships of interest are the relationship between $x$ and $conv(y)$ and between $y$ and $conv(x)$.
Since RCC-22 does not consider tangential relationships between these regions, we can use the RCC-5 CN and regard OUTSIDE($x,y$) as meaning that \DC($x,conv(y)$), P-INSIDE($x,y$) as meaning \PO($x,conv(y)$) and INSIDE($x,y$) as meaning \P($x,conv(y)$).
Thus a relation of the form P-INSIDE\_$\beta$\_$\gamma$ can transition to OUTSIDE\_$\beta$\_$\gamma$ or  INSIDE\_$\beta$\_$\gamma$ (or vice versa). However,  INSIDE\_$\beta$\_$\gamma$ cannot transition to OUTSIDE$\beta$\_$\gamma$, or vice versa. Similar reasoning applies to $\alpha$\_P-INSIDE\_$\gamma$.

The tricky case is whether a transition of both $\alpha$ and $\beta$ can occur simultaneously.
This computation relies on the concept of \emph{dominance} \citep{galton2000qualitative}, which captures the notion that some relations may only hold instantaneously (e.g. \EQ, \EC\  and \TPP) whilst others must hold over an interval (e.g. \PO, \NTPP). If two relations hold simultaneously and one dominates the other, then they cannot both transition simultaneously -- the former will transition first. Since, as explained just above, the CN underlying the transition between the various \EC\ and \DC\  cases is the RCC-5 CN, and thus the  dominance of  \PO\ (which relates to P-INSIDE as explained above), \DR\ (which relates to OUTSIDE) and \PP\  (which relates to the INSIDE relation). This is made explicit in the dominates/2 relation defined immediately below.  Since \DR\  includes the RCC-8 relation \EC, \DR\  dominates \PO\  and thus OUTSIDE dominates P-INSIDE. Similarly, since the RCC-5 relation \PP\  includes the RCC-8 relation \TPP, INSIDE dominates P-INSIDE.

Note that in any CN transition, one of the nodes will always dominate the other; even in the case for RCC-8 CN, where, e.g.  \TPP\ dominates \PO, and \EQ\ dominates \PO, \EQ\  dominates \TPP\  -- reflecting the fact that it is impossible to move from a \TPP\   situation to an \EQ\  in an instant, whereas the reverse is possible instantaneously.

\begin{lstlisting}[basicstyle=\ttfamily\scriptsize, breaklines=true]

%define dominates/2 to specify which relations dominate which other ones
dominates(o,p). %o is short for outside and p short for partially-inside
dominates(i,p). %i is short for inside
dominates(e,d). %e is short for EC and d for DC
dominates(X,X).  % this is to allow a relation to stay the same;
        % strictly, this is an abuse of  notation:
        % dominates is an irreflexive relation  --
        % a relation does not dominate itself,
        % but it is convenient to allow one component to stay the same
        % A special check to ensure at least one relation
        %changes (the \=  condition in the edge/2 relation below.

%define rel/1  defines which relations can instatiate the first two components of
    %an alpha_beta_gamma relation (e.g. I_O_E)
%define drel to name the two possibilities for the third component (gamma)
%These 2 predicates are used to  instantiate the variables at the
    %start of the clause.
%Could have also done this by using dominates/2 to instantiate and having
    %two versions of dominates/2.

rel(o).
rel(i).
rel(p).
drel(e).
drel(d).

%edge/2 is the core predicate which computes the edges in the CN;
%one for going from dominating relations to non dominating and one
%from going the other way round.  Note that we can only have one kind
%of transition in any edge as dominating ones will happen before any
%non dominating ones.
%The \= conditions in each clause ensure that the two relations are not
%identical, and that the impossible relations IIE and IID are excluded.

edge([X1,Y1,Z1],[X2,Y2,Z2]) :- rel(X1), rel(Y1), drel(Z1), dominates(X1,X2), dominates(Y1,Y2), dominates(Z1,Z2), [X1,Y1,Z1] \= [X2,Y2,Z2], [X1,Y1] \= [i,i], [X2,Y2] \= [i,i].

edge([X1,Y1,Z1],[X2,Y2,Z2]) :- rel(X1), rel(Y1), drel(Z1), dominates(X2,X1), dominates(Y2,Y1), dominates(Z2,Z1), [X1,Y1,Z1] \= [X2,Y2,Z2], [X1,Y1] \=  [i,i], [X2,Y2] \= [i,i].


%predicate to collect all the edges into a list L
alledges(L) :- setof([X,Y], edge(X,Y), L).
%can the print these to a file called, e.g., rcc22cn by querying
%alledges(L), print_pairs_to_file(L,rcc22cn).

% print_pairs_to_file(List, Filename)
% Writes each [X,Y] element of List to Filename in format X>Y.

print_pairs_to_file(List, Filename) :-
    open(Filename, write, Stream),
    write_pairs(Stream, List),
    close(Stream).

% Helper predicate to write each pair in the specified format.
write_pairs(_, []).
write_pairs(Stream, [[X, Y]|Tail]) :-
    format(Stream, '~w > ~w~n', [X, Y]),
    write_pairs(Stream, Tail).


\end{lstlisting}

\end{document}